\documentclass[lettersize,journal]{IEEEtran}

\usepackage[numbers,sort&compress]{natbib}
\usepackage[pdftex]{graphicx}
\usepackage{amssymb}
\usepackage{amsmath,amsfonts}
\usepackage{amsthm}
\usepackage[ruled,lined,linesnumbered]{algorithm2e}
\usepackage{array}
\usepackage{fixltx2e}
\usepackage[caption=false,font=normalsize]{subfig}
\usepackage{textcomp}
\usepackage{stfloats}
\usepackage{url}
\usepackage{verbatim}
\usepackage{graphicx}
\usepackage{xcolor}
\usepackage{comment}
\usepackage{multirow}
\usepackage{ulem}
\usepackage{dsfont}

\allowdisplaybreaks[2]

\def\BibTeX{{\rm B\kern-.05em{\sc i\kern-.025em b}\kern-.08em
    T\kern-.1667em\lower.7ex\hbox{E}\kern-.125emX}}
\usepackage{balance}

\newtheorem{thm}{Theorem}
\newtheorem{lem}{Lemma}

\newtheorem{assum}{Assumption}
\newtheorem{cor}{Corollary}
\theoremstyle{definition}
\newtheorem{defn}{Definition}
\newtheorem{rem}{Remark}

\usepackage[T1]{fontenc}
\begin{document}

\title{Automated Federated Learning in Mobile Edge Networks --- Fast Adaptation and Convergence}

\author{Chaoqun You,~\IEEEmembership{Member,~IEEE,}
        Kun Guo,~\IEEEmembership{Member,~IEEE,}\\
        Gang Feng,~\IEEEmembership{Senior Member,~IEEE,}\
        Peng Yang,~\IEEEmembership{Member,~IEEE,}\\
        and~Tony~Q.~S.~Quek,~\IEEEmembership{Fellow,~IEEE}
\thanks{C. You and T. Quek are with the Singapore University of Design and Technology, 487372, Singapore (e-mail: chaoqun\_you@sutd.edu.sg and tonyquek@sutd.edu.sg).}
\thanks{K. Guo is with the East China Normal University, Shanghai 200241, P.R.China (e-mail:kguo@cee.ecnu.edu.cn).}
\thanks{G. Feng is with University of Electronic Science and Technology of China, Chengdu 611731, P.R.China (e-mail:fenggang@uestc.edu.cn).}
\thanks{P. Yang is with the Beihang University, Beijing 100088, P.R.China (e-mail:yangp09@buaa.edu.cn).}}

\markboth{Journal of \LaTeX\ Class Files,~Vol.~14, No.~8, August~2021}%
{Shell \MakeLowercase{\textit{et al.}}: A Sample Article Using IEEEtran.cls for IEEE Journals}


\maketitle

\begin{abstract}
Federated Learning (FL) can be used in mobile edge networks to train machine learning models in a distributed manner. Recently, FL has been interpreted within a Model-Agnostic Meta-Learning (MAML) framework, which brings FL significant advantages in fast adaptation and convergence over heterogeneous datasets. However, existing research simply combines MAML and FL without explicitly addressing how much benefit MAML brings to FL and how to maximize such benefit over mobile edge networks. In this paper, we quantify the benefit from two aspects: optimizing FL hyperparameters (i.e., sampled data size and the number of communication rounds) and resource allocation (i.e., transmit power) in mobile edge networks. Specifically, we formulate the MAML-based FL design as an overall learning time minimization problem, under the constraints of model accuracy and energy consumption. Facilitated by the convergence analysis of MAML-based FL, we decompose the formulated problem and then solve it using analytical solutions and the coordinate descent method. With the obtained FL hyperparameters and resource allocation, we design a MAML-based FL algorithm, called Automated Federated Learning (AutoFL), that is able to conduct fast adaptation and convergence. Extensive experimental results verify that AutoFL outperforms other benchmark algorithms regarding the learning time and convergence performance.
\end{abstract}

\begin{IEEEkeywords}
Fast adaptation and convergence, federated learning, model-agnostic meta-learning, mobile edge networks
\end{IEEEkeywords}

\section{Introduction} \label{sec:1}

\IEEEPARstart{I}{n} recent years, modern mobile user equipments (UEs) such as smart phones and wearable devices have been equipped with advanced powerful sensing and computing capabilities~\cite{ghahramani2020urban}.
This enables their access to a wealth of data that are suitable for learning models, which brings countless opportunities for meaningful applications such as Artificial Intelligence (AI) medical diagnosis~\cite{pryss2015mobile} and air quality monitoring~\cite{ganti2011mobile}.
Traditionally, learning models requires data to be processed in a cloud data center~\cite{li2017multi}.
However, due to the long distance between the devices where data is generated and the servers in data centers, cloud-based Machine Learning (ML) for mobile UEs may incur unacceptable latencies and communication overhead.
Therefore, Mobile Edge Computing (MEC)~\cite{mao2017survey, wang2020convergence} has been proposed to facilitate the deployment of servers near the base station (BS) at mobile edge networks, so as to bring intelligence to network edge.

In mobile edge networks, for a traditional ML paradigm, it is inevitable to upload raw data from mobile UEs to a server which could be deployed near the BS for model learning.
However, the unprecedented amount of data created by mobile UEs are private in nature, leading to the increasing concern of data security and user privacy.
To address this issue, Federated Learning (FL) has been proposed~\cite{mcmahan2017communication} as a new ML paradigm.
FL in mobile edge networks refers to training models across multiple distributed UEs without ever uploading their raw data to the server.
In particular, UEs compute local updates to the current global model based on their own local data, which are then aggregated and fed-back by an edge server, so that all UEs have access to the same global model for their new local updates.
Such a procedure is implemented in one communication round and is repeated until a certain model accuracy is reached.
%

Despite its promising benefits, FL also comes with new challenges in practice. Particularly, the datasets across UEs are \textit{heterogenous}.
Not only does the number of data samples generated by UEs vary, but these data samples are usually not independent and identically distributed (non-i.i.d).
Learning from such heterogeneous data is not easy, as conventional FL algorithms usually develop a \textit{common} model for all UEs [8–10] such that, the global model obtained by minimizing the average loss could perform arbitrarily poorly once applied to the local dataset of a specific UE.
That is, the global model derived from conventional FL algorithms may conduct weak adaptations to local UEs. Such weak adaptations will further restrict the convergence rate of the FL algorithms.

Multiple techniques are emerging as promising solutions to the data heterogeneity problem, such like adding user context~\cite{mansour2020three}, transfer-learning~\cite{schneider2019mass}, multi-task learning~\cite{smith2017federated}, and Model-Agnostic Meta-Learning (MAML)~\cite{pmlr-v70-finn17a}. Of all these techniques, MAML is the only one that not only addresses the heterogenous data problem, but greatly speeds up the FL learning process as well.
MAML is a well-known meta-learning~\cite{vanschoren2019meta} approach that learns from the past experience to adapt to new tasks much faster.
Such adaptations make it possible for MAML to develop well-behaved user-specific models.
More specifically, MAML aims to find a sensitive \textit{initial} point that learned from the past experience to conduct \textit{fast} adaptations requiring only a few data points on each UE.
With the learned initial model, MAML dramatically speeds up the learning process by replacing hand-engineered algorithms with an \textit{automated}, data-driven approach~\cite{hutter2019automated}.
Fortunately, in both MAML and FL, existing algorithms use a variant of gradient descent method locally, and send an overall update to a coordinator to update the global model.
This similarity makes it possible to interpret FL within a MAML framework~ \cite{fallah2020personalized,jiang2019improving,deng2020adaptive,dinh2020personalized}.
Such a simple MAML-based FL is termed as Personalized Federated Learning (PFL), and the algorithm that realizes PFL is termed as Personalized FederatedAveraging (Per-FedAvg).
However, Per-FedAvg is a pure ML algorithm, and how to apply it into practical mobile edge networks remains \textit{unclear}.
Recently several attempts have been made to study the implementation of Per-FedAvg in practice. \cite{wu2020personalized} proposes a framework to execute Per-FedAvg for intelligent IoT applications without considering Per-FedAvg's strength in saving learning time. \cite{yue2021inexact}, although aims to achieve fast learning for IoT applications at the edge, does not consider the resource allocation in a practical mobile edge network.
Therefore, to what degree a MAML-based FL algorithm expedites FL in mobile edge networks and under what conditions such benefit could be achieved still remain to be unexplored areas.

In order to quantify the benefit MAML brings to FL in mobile edge networks, we consider two aspects to minimize the overall learning time: optimizing learning-related hyperparameters as well as resource allocation.
On the one hand, the typical FL hyperparameters, including the sampled data sizes across UEs and the number of communication rounds, have significant impact on the overall learning time and thus, these hyperparameters need to be carefully specified.
On the other hand, the resource allocation should be considered to account for UEs' practical wireless communication environments and limited battery lifetimes. 
Particularly, due to the existence of random channel gain and noise over wireless channels, the transmit power allocation on an UE will decide whether the transmitted updates from the UE can be successfully decoded by the edge server. Restricted by the limited battery lifetime, it will also determine whether the remaining energy on that UE is sufficient to support its local training.

It is non-trivial to solve the formulated optimization problem considering above two quantitative dimensions, given that the relationship between the variables (i.e., the sampled data sizes across UEs, the number of communication rounds, and the transmit power on UEs) and the model accuracy $\epsilon$ is implicit.
Therefore, we start with the convergence analysis of the MAML-based FL algorithm, by which the three variables are bounded as functions of $\epsilon$.
After that, the formulated optimization problem can be approximatively decoupled into three sub-problems, each of which accounts for one of the three variables. Specifically, solving the first sub-problem gives an insight to achieve the required number of communication round.
As for the other two sub-problems, we use the coordinate descent method~\cite{wright2015coordinate} to compute the sampled data size and the transmit power of all UEs iteratively.
In other words, we first give an initial value of the transmit power of UEs, and use it to compute the sampled data sizes in the second sub-problem; then with the attained sampled data sizes, we compute the transmit power in the third sub-problem. This process repeats until a certain model accuracy is achieved.

The solution to the optimization problem guides us to the design of Automated Federated Learning (AutoFL).
AutoFL uses the results derived from the optimization problem to design the learning process, thereby quantifying as well as maximizing the benefit MAML brings to FL over mobile edge networks.
More specifically, in each round $k$, the BS first sends the current model parameter $w_k$ to a random subset of UEs.
According to the given model accuracy $\epsilon$, the sampled data sizes and the transmit power of the selected UEs are determined by our proposed solution. Each selected UE then trains its local model with the determined sampled data size and then transmits the local update to the edge server with the determined transmit power.
The server receives local updates from the selected UEs and decides whether these updates can be successfully decoded.
Those successfully decoded updates are then aggregated to update the global model as $w_{k+1}$. Such a communication round is executed repeatedly until the model accuracy $\epsilon$ is achieved. In this way, AutoFL is able to inherit the advantages of MAML over mobile edge networks, thereby conducting model learning with fast adaptation and convergence.

To summarize, in this paper we make the following contributions:
\begin{itemize}
  \item We provide a comprehensive problem formulation for the MAML-based FL design over mobile edge networks, which can account for the impact of FL hyperparameters and network parameters on the model accuracy and learning time at the same time. Specifically, we jointly optimize the UEs' sampled data size and transmit power, as well as the number of communication rounds, to quantify and maximize the benefit MAML brings to FL, by which the overall learning time is minimized under the constraint of model accuracy $\epsilon$ and energy consumption.
  \item We analyse the convergence of MAML-based FL algorithm as the first step to solve the formulated problem. The convergence analysis makes the relationships between optimization variables and the model accuracy explicit, especially characterizes the sampled data size, the transmit power, and the number of communication rounds as functions of model accuracy $\epsilon$.
  \item Applying the obtained results of convergence analysis, the formulated problem is decoupled into three sub-problems. By solving these sub-problems, the number of communication rounds can be characterized using a closed-form solution, while the sampled data sizes and the transmit power of all UEs in each round can be achieved using the coordinate descent method. With the optimized hyperparameters and resource allocation, we further propose AutoFL with fast adaptation and convergence.
  \item By conducting extensive experiments with MNIST and CIFAR-10 datasets, we demonstrate the effectiveness and advantages of AutoFL over Per-FedAvg and FedAvg, two baseline FL algorithms in terms of learning time, model accuracy and training loss.
\end{itemize}

The rest of this paper is organized as follows. We first give the system model and problem formulation in Section \ref{sec:2}. Then we give the convergence analysis to make the formulated problem tractable in Section \ref{sec:3}. We recast the optimization problem and then propose the solutions to guide the design of AutoFL in Section \ref{sec:4}. Extensive experimental results are presented and discussed in Section \ref{sec:5}. Finally, we conclude this paper in Section \ref{sec:6}.

\section{System Model and Problem Formulation} \label{sec:2}

\begin{figure}[!t]
\centering
  \includegraphics[width=3in]{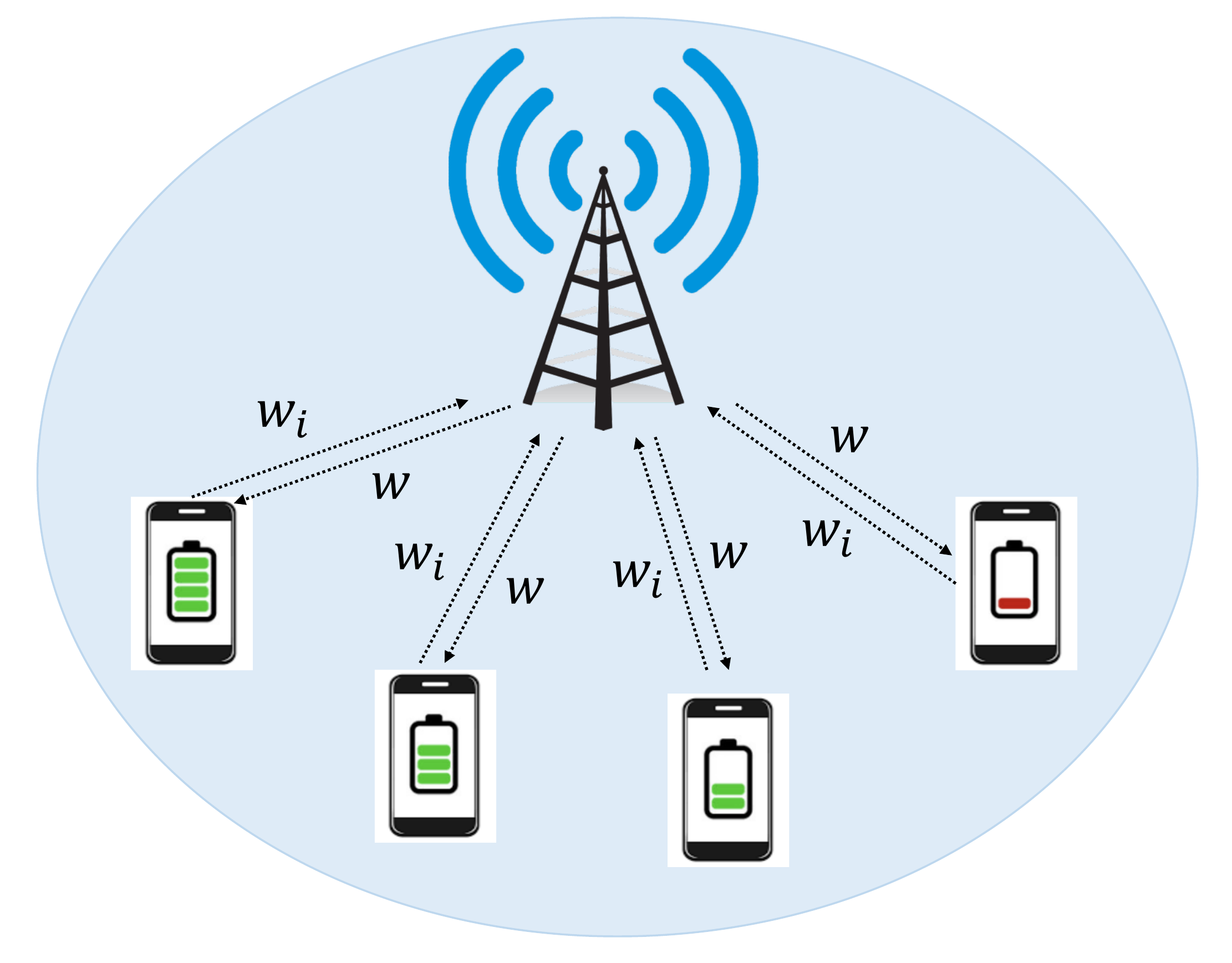}
  \caption{An illustration of mobile edge network.}
  \label{fig_1}
\end{figure}

Consider a typical mobile edge network with a edge server co-deployed with a BS and $n$ UEs, and the UEs are indexed by $\mathcal{U} = \{1,\dots, n\}$, as illustrated in Fig.~\ref{fig_1}.
In this section, we first explain why the FL problem can be interpreted within the MAML framework. Based on this framework, we introduce our system model and problem formulation.

\subsection{Interpreting FL within the MAML framework} \label{sec:2.1}

In FL, we consider a set of $n$ UEs which are connected with the server via the BS, where each UE has access only to its local data~\cite{mcmahan2017communication}.
For a sample data point $\{x,y\}$ with input $x$, the goal of the server is to find the model parameter $w$ that characterizes the output $y$ with loss function $f(w):\mathbb{R}^m \rightarrow \mathbb{R}$, such that the value of $f(w)$ can be minimized.
More specifically, if we define $f_i(w):\mathbb{R}^m \rightarrow \mathbb{R}$ as the loss function of UE $i$, the goal of the server is to solve
\begin{equation} \label{equ:globalFL}
  \min_{w\in \mathbb{R}^m} f(w):=\frac{1}{n} \sum_{i=1}^n f_i(w).
\end{equation}
In particular, for each UE $i$, we have
\begin{equation} \label{equ:localFL}
  f_i(w):= \frac{1}{D_i} \sum_{(x,y)\in\mathcal{D}_i} l_i(w;x,y),
\end{equation}
where $l_i(w;x,y)$ is the error between the true label $y\in\mathcal{Y}_i$ and the prediction of model $w$ using input $x \in\mathcal{X}_i$. Each UE $i$ has a local dataset $\mathcal{D}_i = \{x\in\mathcal{X}_i, y\in\mathcal{Y}_i\}$, with $D_i = |\mathcal{D}_i|$ data samples. Since the datasets captured by the UEs are naturally \textit{heterogenous}, the probability distribution of $\mathcal{D}_i$ across UEs is not identical.

MAML, on the other hand, is one of the most attractive technique in meta-learning.
MAML is proposed to learn an \textit{initial} model that adapts quickly to a new task through one or more gradient steps with only a few data points from that new task.
Each task can be regarded as an object with its own dataset to learn, just like an UE in FL.
MAML allows us to replace hand-engineered algorithms with data-driven approaches to learn initial parameters \textit{automatically}.

To show how to exploit the fundamental idea behind the MAML framework~\cite{pmlr-v70-finn17a} to design an automated variant of FL algorithm, let us first briefly recap the MAML formulation.
In MAML, if we regard the tasks as UEs and assume each UE takes the initial model and updates it using one step of gradient descent with respect to its own loss function\footnote{A more general case is to perform multiple steps of gradient descent. However, this would lead to expensive cost of computing multiple Hessians. Therefore, for simplicity, throughout the whole paper we consider only one single step of gradient descent.}, problem (\ref{equ:globalFL}) then changes to
\begin{equation} \label{equ:globalMAML}
  \min_{w\in \mathbb{R}^m} F(w):= \frac{1}{n} \sum_{i=1}^n f_i(w-\alpha \nabla f_i(w)),
\end{equation}
where $\alpha\geq0$ is the learning rate at a UE. For UE $i$, its optimization objective $F_i(w)$ can be expressed as
\begin{equation} \label{equ:localMAML}
  F_i(w):= f_i(w-\alpha\nabla f_i(w)).
\end{equation}

Such a transformation from problem (\ref{equ:globalFL}) to (\ref{equ:globalMAML}) implies that the FL problem can be interpreted within the MAML framework.
The FL algorithm proposed to solve (\ref{equ:globalMAML}) is termed as Personalized Federated Averaging (Per-FedAvg)~\cite{fallah2020personalizedArxiv, fallah2020personalized}.
Per-FedAvg is inspired by FedAvg, which is proposed in~\cite{mcmahan2017communication} as a classic and general FL algorithm to solve (\ref{equ:globalFL}) in a distributed manner.

Per-FedAvg is summarized in Algorithm~\ref{alg:Per-FedAvg}.
In each round $k$ $(k=1,\dots,K)$, the central BS randomly picks a set of UEs $\mathcal{A}_k$ and then sends them the current model parameter $w_k$.
Each UE first adapts the global model $w_k$ to its local data and obtains an intermediate parameter $\theta_k^i$, where $\theta_k^i = w_k- \alpha \nabla f_i(w_k)$. Then, with $\theta_k^i$, UE $i$ updates its local model using one or more steps of gradient descent and obtains $w_{k+1}^i$. Such local model parameter updates are then sent to the server for model aggregation. The server will update the global model or meta model as $w_{k+1}$.
%
This process repeats in the sequel rounds until a certain model accuracy is achieved or a predefined maximum number of rounds is reached.
\begin{algorithm}[h]
    \caption{Per-FedAvg Pseudo-code}
    \label{alg:Per-FedAvg}
    \SetKwInOut{Input}{Input}
    \Input{\begin{minipage}[t]{0.9\linewidth}
      UE learning rate: $\alpha$; \\
      BS learning rate: $\beta$; \\
      Initial model parameter: $w_0$;
    \end{minipage}}
    \For{$k=1$ \KwTo $K$}{
        Choose a subset of UEs $\mathcal{A}_k$ uniformly at random with size $A_k$\;
        BS sends $w_k$ to all UEs in $\mathcal{A}_k$\;
        \For{each UE $i \in \mathcal{A}_k$}{
            Compute $\theta_k^i := w_k - \alpha \nabla_{w_k} f_i(w_k)$ \;
            Compute $w_{k+1}^i := w_k - \beta \nabla_{w_k} F_i(\theta_k^i) = w_k - \beta (I-\nabla_{w_k}^2 f_i(w_k))\nabla_{\theta_k^i} f_i(\theta_k^i)$\;
            UE $i$ sends $w_{k+1}^i$ to the central BS.
        }
        BS updates the global model as: \
        $w_{k+1} = \frac{1}{A_k} \sum_{i\in \mathcal{A}_k} w_{k+1}^i$\;
    }
\end{algorithm}

Per-FedAvg is a MAML-based FL algorithm, which suggests a general approach to use the MAML method for solving the data heterogeneity problem in FL.
It is proposed with focus on the personalization in FL, which is a natural feature inherited from MAML.
Except for the personalization, MAML is also a few-shot learning approach, as only a few samples are required to learn new skills after just minutes of experience.
Therefore, Per-FedAvg also inherits this feature of MAML, thereby adapting quickly from only a few samples.
To what degree of fast adaptation MAML benefits FL and under what conditions such benefit can be achieved is what exactly this paper is about.

\subsection{Machine Learning Model} \label{sec:2.2}

We consider the above described MAML-based FL. In detail, we concentrate on the situation where UEs communicate in a \textit{synchronous} manner, so as to avoid using outdated parameters for global model update and make high-quality refinement in each round.
Meanwhile, for each UE, we consider the case that only one step of stochastic gradient descent (SGD) is performed, following the same setting as~\cite{pmlr-v70-finn17a}.

As for the concerned MAML-based FL, our goal is to optimize the initial model using only a few data points at each UE. Hence, we only obtain an estimate of the desired gradient with SGD.
Here, the desired gradient $\nabla F_i(w)$ on UE $i$ is computed using all data points in its dataset $\mathcal{D}_i$, while the estimated gradient $\tilde{\nabla} F_i(w)$ on UE $i$ is computed using SGD with the sampled dataset $\mathcal{D}_i^{(\cdot)} \in \mathcal{D}_i$. Note that, the superscript $(\cdot)$ represents different sampled datasets are used to estimate the involved gradients and Hessian in $\nabla F_i(w)$. Meanwhile, the sampled data size is $D_i^{(\cdot)}=|\mathcal{D}_i^{(\cdot)}|$.

More specifically, in order to solve (\ref{equ:globalMAML}), each UE $i$ computes the desired gradient in round $k$, as follows:
\begin{equation}\label{equ:nabla_F}
  \nabla F_i(w_k) = (I - \alpha \nabla^2 f_i(w_k)) \nabla f_i(w_k-\alpha \nabla f_i(w_k)).
\end{equation}
At every round, computing the gradient $\nabla f_i(w_k)$ by using all data points of UE $i$ is often computationally expensive. Therefore, we take a subset $\mathcal{D}_i^{\text{in}}$ from $\mathcal{D}_i$ to obtain an unbiased estimate $\tilde{\nabla} f_i(w_k;\mathcal{D}_i^{\text{in}})$ for $\nabla f_i(w_k)$, which is given by
\begin{equation}\label{equ:tilde_nabla_f}
  \tilde{\nabla} f_i(w_k;\mathcal{D}_i^{\text{in}})=\frac{1}{D_i^{\text{in}}} \sum_{(x,y)\in \mathcal{D}_i^{\text{in}}} \nabla l_i(w_k;x,y).
\end{equation}
Similarly, the outer gradient update $\nabla f_i(\theta_k^i)$ and Hessian update $\nabla^2 f_i(w_k)$ in (\ref{equ:nabla_F}) can be replaced by their unbiased estimates $\tilde{\nabla} f_i(\theta^i_k;\mathcal{D}_i^{\text{o}})$ and $\tilde{\nabla}^2 f_i(w_k;\mathcal{D}_i^{\text{h}})$ respectively. Here, $\mathcal{D}_i^{\text{o}}$ and $\mathcal{D}_i^{\text{h}}$ are sampled from $\mathcal{D}_i$ as well.
Therefore, using SGD, we can finally obtain an estimated local gradient $\tilde{\nabla} F_i(w_k)$ on UE $i$ in round $k$, which is given by
\begin{equation}\label{equ:local_update_estimate}
  \tilde{\nabla} F_i(w_k) = (I - \alpha \tilde{\nabla}^2f_i(w_k;\mathcal{D}_i^{\text{h}})) \tilde{\nabla} f_i(w_k-\alpha\tilde{\nabla}f_i(w_k;\mathcal{D}_i^{\text{in}}); \mathcal{D}_i^{\text{o}}).
\end{equation}

It is worth noting that $\tilde{\nabla} F_i(w_k)$ is a biased estimator of $\nabla F_i(w_k)$. This is because the stochastic gradient
$\tilde{\nabla} f_i(w_k -\alpha \tilde{\nabla} f_i(w_k;\mathcal{D}_i^{\text{in}});\mathcal{D}_i^{\text{o}})$ contains another stochastic gradient $\tilde{\nabla}f_i(w_k;\mathcal{D}_i^{\text{in}})$ inside.
Hence, to improve the estimate accuracy, $\mathcal{D}_i^{\text{in}}$ used for inner gradient update is independent from the sampled datasets $\mathcal{D}_i^{\text{o}}$ and $\mathcal{D}_i^{\text{h}}$ used for outer gradient and Hessian update respectively.
Meanwhile, in this paper we assume $\mathcal{D}_i^{\text{o}}$ and $\mathcal{D}_i^{\text{h}}$ are also independent from each other.

\subsection{Communication and Computation Model}

When exploring the MAML-based FL in realistic mobile edge networks, the communication and computation model should be captured carefully. Particularly, we consider that UEs access the BS through a channel partitioning scheme, such as orthogonal frequency division multiple access (OFDMA).
In order to successfully upload the local update to the BS, two conditions need to be satisfied: (1) the UE is selected, and (2) the transmitted local update is successfully decoded. In this respect, we first introduce $s_i^k \in \{0,1\}$ as a selection indicator, where $s_k^i=1$ indicates the event that UE $i$ is chosen in round $k$, and $s_k^i=0$ otherwise. Next, we characterize the transmission quality of the wireless links. For the signals transmitted from UE $i$, the SNR received at the BS can be expressed as $\xi_k^i = \frac{p_k^i h_k^i \|c_i\|^{-\kappa}}{N_0}$, where $p_k^i$ is the transmit power of UE $i$ during round $k$, $\kappa$ is the path loss exponent. $h_k^i \|c_i\|^{-\kappa}$ is the channel gain between UE $i$ and the BS with $c_i$ being the distance between UE $i$ and the BS and $h_k^i$ being the small-scale channel coefficient.
$N_0$ is the noise power spectral density. In order for the BS to successfully decode the local update from UE $i$, it is required that the received SNR exceeds a decoding threshold $\phi$, i.e., $\xi_k^i > \phi$.
Assume that the small-scale channel coefficients across communication rounds follow Rayleigh distribution, then according to~\cite{yang2019scheduling}, the update success transmission probability, defined as $q_k^i = \mathbb{P}(s_k^i = 1, \xi_k^i > \phi)$, can be estimated as follows:
\begin{equation} \label{equ:probability}
  q_k^i = \mathbb{P}(s_k^i = 1, \xi_k^i > \phi)\approx \frac{1/n}{1+ \nu(p_k^i)}.
\end{equation}
where $\nu(p_k^i) = \frac{N_0\phi}{p_k^i}$. Then the achievable uplink rate $r_{k,i}$ of UE $i$ transmitting its local update to the BS in round $k$ is given by
\begin{equation}
  r_{k,i} = B\log_2(1+\xi_k^i),
\end{equation}
where $B$ is the bandwidth allocated to each UE. Based on $r_{k,i}$, the uplink transmission delay of UE $i$ in round $k$ can be specified as follows:
\begin{equation}
  t_{k,i}^{\text{com}} = \frac{Z}{r_{k,i}},
\end{equation}
where $Z$ is the size of $w_k$ in number of bits. Since the transmit power of the BS is much higher than that of the UEs, the downlink transmission delay is much smaller than the uplink transmission delay. Meanwhile, we care more about the transmit power allocation on individual UEs rather than that on the BS, so here we ignore the downlink transmission delay for simplicity.


Further, we calculate the computation time for each UE, which is consumed for computing the local update. Given the CPU-cycle frequency of UE $i$ by $\vartheta_i$, the computation time of UE $i$ is expressed as
\begin{equation}
  t_{k,i}^{\text{cmp}} = \frac{c_i d_k^i}{\vartheta_i}.
  \label{eq:cmp_time}
\end{equation}
In (\ref{eq:cmp_time}), $c_i$ denotes the number of CPU cycles for UE $i$ to execute one sample of data points and $d_k^i= D_{k,i}^{\text{in}} + D_{k,i}^{\text{o}} + D_{k,i}^{\text{h}}$ denotes the sampled data size of UE $i$ in round $k$. 

In term of $t_{k,i}^{\text{com}}$ and $t_{k,i}^{\text{cmp}}$, we then give the energy consumption of each UE $i$ in round $k$, which consists of two parts: (1) the energy for transmitting the local updates and (2) the energy for computing the local updates. Let $e_k^i$ denote the energy consumption of each UE $i$ in round $k$, and then $e_k^i$ can be computed as follows~\cite{tran2019federated, pei2020energy}:
\begin{equation}
  e_k^i =  \frac{\varsigma}{2} \vartheta_i^3t_{k,i}^{\text{cmp}} + p_k^i t_{k,i}^{\text{com}},
  \label{eq:energy_UE}
\end{equation}
where $\frac{\varsigma}{2}$ is the effective capacitance coefficient of UE $i$'s computing chipset. From (\ref{eq:energy_UE}), we observe that in round $k$ for each UE $i$, both the sampled data size $d_i^k$ and transmit power $p_i^k$ have significant impacts on its energy consumption. This observation motivates us to jointly consider these two variables when quantifying the benefit MAML brings to FL in mobile edge networks.

\subsection{Problem Formulation} \label{sec:2.3}

In order to quantify the benefit MAML brings to FL in mobile edge networks, we focus on the learning time minimization under the constraint of model accuracy and UEs' energy consumption. Particularly, the learning time is the duration over all $K$ communication rounds, in which the duration of round $k$ is determined by the slowest UE as follows,
\begin{equation}
  T_k^{\text{round}} = \max_{i\in \mathcal{A}_k}\{t_{k,i}^{\text{cmp}} + t_{k,i}^{\text{com}}\}.
\end{equation}
Note that we can replace $i\in\mathcal{A}_k$ with $i\in\mathcal{U}$ in $T_k^{\text{round}}$, that is, $T_k^{\text{round}} = \max_{i\in\mathcal{U}} \{t_{k,i}^{\text{cmp}} + t_{k,i}^{\text{com}}\}$. This is because the UEs that are not chosen in $\mathcal{A}_k$ has a $T_k^i$ with the value of $0$, which would not effect the result of $T_k^{\text{round}}$.
Let $\mathbf{d} \triangleq \{\mathbf{d}^1,\dots,\mathbf{d}^n\}$ and $\mathbf{p} \triangleq \{\mathbf{p}^1,\dots,\mathbf{p}^n\}$ denote the UEs' sampled data size vector and the transmit power vector respectively, where $\mathbf{d}^i \triangleq \{d_1^i,\dots,d_K^i\}$ and $\mathbf{p}^i \triangleq \{p_1^i,\dots,p_K^i\}$.
Then, we can give the problem formulation by
\begin{subequations}\label{equ:opt1}
  \begin{align}
    \min_{\mathbf{d},\mathbf{p},K} \qquad & \sum_{k=0}^{K-1} \max_{i\in\mathcal{U}} \{ t_{k,i}^{\text{cmp}} + t_{k,i}^{\text{com}}\} \label{equ:opt1_obj} \tag{P1}\\
    \text{s.t.} \qquad & \frac{1}{K}\sum_{k=0}^{K-1} \mathbb{E}[\|\nabla F(w_k)\|^2] \leq \epsilon, \label{equ:opt1.1} \tag{C1.1} \\
             & 0 \leq e_k^i \leq E_{\max},\quad \forall i \in\mathcal{U} \label{equ:opt1.2},k=0,...,K-1  \tag{C1.2}\\
             & 0 \leq p_k^i \leq P_{\max}, \quad \forall i\in \mathcal{U}, k=0,...,K-1 \label{equ:opt1.3} \tag{C1.3}\\
             & 0 \leq d_k^i \leq D_i,\quad \forall i\in\mathcal{U}, k=0,...,K-1. \label{equ:opt1.4} \tag{C1.4}
  \end{align}
\end{subequations}

In problem (P1), we not only optimize $\mathbf{d}$ and $\mathbf{p}$, but also the total number of rounds $K$. The hidden reason is that, both $\mathbf{d}$ and $\mathbf{p}$ have an effect on the duration of each round, thereby impacting the number of rounds $K$ to achieve a certain model accuracy $\epsilon$. Besides, (\ref{equ:opt1.1}) characterizes an $\epsilon$-approximate convergence performance.
(\ref{equ:opt1.2}) limits the energy consumption of UE $i$ in round $k$ not larger the predefined maximum value $E_{\max}$.
(\ref{equ:opt1.3}) gives the maximum transmit power of UE $i$ in round $k$ by $P_{\max}$.
(\ref{equ:opt1.4}) is the sampled data size constraint, that the sampled data size of one UE in round $k$ is smaller than or equal to the data points generated by that UE. 
The solution to problem (P1) can be exploited for designing an optimized MAML-based FL algorithm, which is implemented iteratively to update the global model parameter $w$.
According to the communication and computation model, only those local updates from the selected UEs as well as being successfully decoded by the BS can contribute to updating the global model parameter. That is, in round $k$, we have
\begin{align}\label{equ:update}
  w_{k+1} = & \frac{1}{\sum_{i=1}^{n} \mathds{1}\{s_{k}^i = 1, \xi_k^i>\phi\}} \nonumber \\
  & \sum_{i=1}^{n} \mathds{1}\{s_{k}^i = 1, \xi_k^i\geq\phi\} (w_k - \beta \nabla_{w_k} F(\theta_k^i)),
\end{align}
where $\mathds{1}\{s_{k}^i = 1, \gamma_k^i>\phi\} = 1$ if the event of $s_{k}^i = 1, \gamma_k^i>\phi$ is true, and it equals to zero otherwise.

Based on this update rule, (\ref{equ:opt1.1}) related to the MAML-based FL can be analytically analysed to make the relationship between the decision variables and the model accuracy $\epsilon$ explicit, thereby facilitating solving problem (P1). In this regard, the convergence analysis of the MAML-based FL is given in the following section.

\section{Convergence Analysis of MAML-based FL} \label{sec:3}


\subsection{Preliminaries} \label{sec:3.1}

For a general purpose, we concentrate on the \textit{non-convex} settings on loss functions and aim to find an $\epsilon$-approximate first-order stationary point (FOSP) for the loss function minimization problem (\ref{equ:globalMAML}) in the MAML-based FL, which is defined as follows.
\begin{defn}\label{defn:1}
  A random vector $w_\epsilon\in \mathbb{R}^m$ is called an $\epsilon$-FOSP if it satisfies $\mathbb{E}[\|\nabla F(w_\epsilon)\|^2] \leq \epsilon$.
\end{defn}

In terms of this Definition, we then elaborate on the assumptions, convergence bounds, and discussions on the convergence analysis, respectively, from which we can reformulate (C1.1) in problem (P1) as an explicit constraint with respect to the optimization variables.

Without loss of generality, the assumptions used for the convergence analysis of MAML-based FL algorithm is consistent with that of Per-FedAvg~\cite{fallah2020personalizedArxiv, fallah2020personalized} and are given in the following.

\begin{assum}\label{assum:2}
For every UE $i\in\{1,\dots,n\}$, $f_i(w)$ is twice continuously differentiable. Its gradient $\nabla f_i(w)$ is $L$-Lipschitz continuous, that is,
\begin{equation}\label{equ:assum_2}
  \|\nabla f_i(w)-\nabla f_i(u)\| \leq L\|w-u\|, \qquad \forall w,u\in \mathbb{R}^m.
\end{equation}
\end{assum}

\begin{assum}\label{assum:3}
  For every UE $i\in\{1,\dots,n\}$, the Hessian matrix of $f_i(w)$ is $\rho$-Lipschitz continuous, that is,
  \begin{equation}\label{equ:assum_3}
    \|\nabla^2 f_i(w) - \nabla^2 f_i(u)\| \leq \rho \|w-u\|, \qquad \forall w,u\in\mathbb{R}^m.
  \end{equation}
\end{assum}

\begin{assum}\label{assum:4}
For any $w\in \mathbb{R}^m$, $\nabla l_i(w;x,y)$ and $\nabla^2 l_i(w;x,y)$, computed \textit{w.r.t.} a single data point $(x,y)\in \mathcal{X}_i\times\mathcal{Y}_i$, have bounded variance, that is,
\begin{align}\label{equ:assum_4}
  \mathbb{E}_{(x,y)\thicksim p_i}[\|\nabla l_i(w;x,y)-\nabla f_i(w)\|^2] & \leq \sigma^2_G, \nonumber \\
  \mathbb{E}_{(x,y)\thicksim p_i}[\|\nabla^2l_i(w;x,y)-\nabla^2 f_i(w)\|^2]& \leq \sigma^2_H.
\end{align}
\end{assum}

\begin{assum}\label{assum:5}
  For any $w\in \mathbb{R}^m$, the gradient and Hessian matrix of local loss function $f_i(w)$ and the average loss function $f(w) = (1/n) \sum_{i=1}^{n}f_i(w)$ have bound variance, that is
  \begin{align}\label{equ:assum_5}
    \frac{1}{n}\sum_{i=1}^{n}\|\nabla f_i(w) -\nabla f(w)\|^2 & \leq \gamma_G^2, \nonumber \\
    \frac{1}{n}\sum_{i=1}^{n}\|\nabla^2 f_i(w) - \nabla^2 f(w)\|^2 & \leq \gamma_H^2.
  \end{align}
\end{assum}

Before the convergence analysis, we first introduce three lemmas inherited from~\cite{fallah2020personalized, fallah2020personalizedArxiv} quantifying the smoothness of $F_i(w)$ and $F(w)$, the deviation between $\nabla F_i(w)$ and its estimate $\tilde{\nabla} F_i(w)$, and the deviation between $\nabla F_i(w)$ and $\nabla F(w)$, respectively.

\begin{lem} \label{lem:1}
If Assumptions \ref{assum:2}-\ref{assum:4} hold, then $F_i(w)$ is $L_F$-Lipschitz continuous with $L_F:= 4L+\alpha \rho B$. As a result, the average function $F(w) = (1/n)\sum_{i=1}^{n}F_i(w)$ is also $L_F$-Lipschitz continuous.
\end{lem}

\begin{lem}\label{lem:2}
Recall the gradient estimate $\tilde{\nabla} F_i(w)$ shown in (\ref{equ:local_update_estimate}), which is computed using $\mathcal{D}_i^{in}$, $\mathcal{D}_i^{o}$ and $\mathcal{D}_i^{in}$ that are independent sampled datasets with size $D_i^{in}$, $D_i^{in}$ and $D_i^{in}$, respectively. If the conditions in Assumptions \ref{assum:2}-\ref{assum:4} hold, then for any $\alpha \in (0,1/L]$ and $w\in \mathbb{R}^m$, we have
    \begin{align}\label{equ:lem2}
      \left\|\mathbb{E}\left[ \tilde{\nabla}F_i(w) - \nabla F_i(w) \right]\right\| & \leq \frac{2\alpha L \sigma_G}{\sqrt{D^{in}}}, \\
      \mathbb{E}\left[\|\tilde{\nabla}F_i(w) - \nabla F_i(w)\|^2\right] & \leq \sigma_F^2,
    \end{align}
    where $\sigma_F^2$ is defined as
    \begin{equation} \label{equ:sigma_F}
      \sigma_F^2:= 12\left[B^2 + \sigma_G^2\left[ \frac{1}{D^{o}} + \frac{(\alpha L)^2}{D^{in}}\right]\right]\left[1+\sigma_H^2 \frac{\alpha^2}{4D^{h}}\right] - 12B^2,
    \end{equation}
with $D^{in} = \max_{i\in \mathcal{U}} D_i^{in}$, $D^{o} = \max_{i\in \mathcal{U}} D_i^{o}$ and $D^{h} = \max_{i\in \mathcal{U}} D_i^{h}$.
\end{lem}

\begin{lem} \label{lem:3}
   Given the loss function $F_i(w)$ shown in (\ref{equ:localMAML}) and  $\alpha \in (0,1/L]$, if the conditions in Assumptions \ref{assum:2},~\ref{assum:3}, and \ref{assum:5} hold, then for any $w\in \mathbb{R}^m$, we have
  \begin{equation}\label{equ:lem3}
    \frac{1}{n} \sum_{i=1}^{n}\|\nabla F_i(w) - \nabla F(w)\|^2 \leq \gamma_F^2,
  \end{equation}
  where $\gamma_F^2$ is defined as
  \begin{equation}\label{equ:gamma_F}
    \gamma_F^2:=3B^2\alpha^2\gamma_H^2 +192\gamma_G^2,
  \end{equation}
  with $\nabla F(w) = (1/n)\sum_{i=1}^{n} \nabla F_i(w)$.
\end{lem}

\subsection{Analysis of Convergence Bound}

Let $U_i = \max_{k=\{1,\dots,K\}} U_k^i$, where $U_k^i=  \frac{\mathbb{P}\{s_k^i =1, \xi_k^i > \phi\}}{\sum_{i=1}^{n} \mathbb{P}\{s_k^i =1, \xi_k^i > \phi\}}= \frac{q_k^i}{\sum_{i=1}^{n} q_k^i}$ denotes the normalized update successful probability of UE $i$ in round $k$. Then, with Lemmas~\ref{lem:1},~\ref{lem:2} and~\ref{lem:3}, the expected convergence result of the MAML-based FL within a general mobile edge network we describe in Section~\ref{sec:2} can now be obtained by the following theorem.

\begin{thm} \label{thm:1}
  Given the transmit power vector $\mathbf{p}$, the sampled data size vector $\mathbf{d}$, the number of communication rounds $K$, and the optimal global loss $F(w_{\epsilon})$, the normalized update successful probability $U_i$, and $\alpha \in (0,1/L]$, then we have the following FOSP conditions,
  \begin{align} \label{equ:thm1}
   & \frac{1}{K} \sum_{k=0}^{K-1} \mathbb{E}[\|\nabla F(\bar{w}_k)\|^2] \leq \frac{4(F(w_0) - F(w_{\epsilon}))}{\beta K} \nonumber \\
   & + \left(\sum_{i=1}^{n} U_i^2\right)\left( \beta L_F \sigma_F^2 + \beta L_F \gamma_F^2 + \frac{\alpha^2L^2\sigma_G^2}{D^{in}}\right),
  \end{align}
  where $\bar{w}_k$ is the average of the local updates $w_k^i$ that are successfully decoded by the BS in round $k$. That is, we have
  \begin{equation}\label{equ:w_bar}
    \bar{w}_{k} = \frac{1}{\sum_{i=1}^{n} \mathds{1}\{s_k^i =1, \xi_k^i > \phi\}} \sum_{i=1}^{n}\mathds{1}\{s_k^i =1, \xi_k^i > \phi\} w_{k}^i.
  \end{equation}
\end{thm}

\begin{IEEEproof}
See the Appendix.
\end{IEEEproof}

Unlike the convergence guarantee of Per-FedAvg in~\cite{fallah2020personalized} which is characterized by $K$ and $\mathbf{d}$, the convergence guarantee obtained from Theorem~\ref{thm:1} is characterized by $K$, $\mathbf{d}$ and $U_i$, where $U_i$ is a function of the transmit power $\mathbf{p}^i$. That is, the convergence bound of the proposed MAML-based FL in our paper is described in terms of $K$, $\mathbf{d}$ and $\mathbf{p}$. Therefore, our convergence analysis can combine the FL hyperparameters as well as the resource allocation in mobile edge networks.

\subsection{Discussions}

From Theorem~\ref{thm:1}, we are able to characterize $K$, $\mathbf{d}$ and $\mathbf{p}$ with respect to the predefined model accuracy $\epsilon$.
%
%
According to Theorem~\ref{thm:1} and (C1.1) in problem (P1), it is desired for the right-hand-side of (\ref{equ:thm1}) to be equal to or smaller than $\epsilon$. Consequently, we further present the following corollary.

\begin{cor}\label{cor:1}
 Suppose the conditions in Theorem~\ref{thm:1} are satisfied. If we set the number of total communication rounds as $K = \mathcal{O}(\frac{1}{\epsilon^3})$, the global learning rate as $\beta = \mathcal{O} (\epsilon^2)$, the number of data samples as $d_i = \mathcal{O}(\frac{1}{\epsilon})$,
 then we find an $\epsilon$-FOSP for the MAML-based FL in problem (P1).
\end{cor}

\begin{IEEEproof}
$K=\mathcal{O}(\frac{1}{\epsilon^3})$ and $\beta = \mathcal{O}(\epsilon^2)$ will make sure that the order of magnitude of the first term on the right-hand-side of (\ref{equ:thm1}) to be equal to $\mathcal{O}(\epsilon)$.

Then we examine the second term.
Since $U_i$ is the normalized update successful probability with the consideration of $n$ UEs, its order of magnitude is defined by $n$ rather than $\epsilon$, i.e., $U_i = \mathcal{O}(\frac{1}{n})$.
Therefore, it is natural to take attentions on the term $\beta L_F \sigma_F^2 + \beta L_F \gamma_F^2 + \frac{\alpha^2 L^2 \sigma_G^2}{D^{\text{in}}}$ for an $\epsilon$-FOSP.
By setting $D_i^{\text{in}} = D_i^{\text{o}} = D_i^{\text{h}} = \mathcal{O}(\frac{1}{\epsilon})$ (i.e., $d_i= \mathcal{O}(\frac{1}{\epsilon})$), $\frac{\alpha^2 L^2 \sigma_G^2}{D^{\text{in}}}$ will dominate the value of $\beta L_F \sigma_F^2 + \beta L_F \gamma_F^2 + \frac{\alpha^2 L^2 \sigma_G^2}{D^{\text{in}}}$.
This is because, in this case, $\beta L_F \sigma_F^2 + \beta L_F \gamma_F^2 = \mathcal{O}(\epsilon^4)$, which is a high-order infinitesimal of $\epsilon$.
Finally, combining the magnitudes of the first and second terms, we can conclude that the joint of $K=\mathcal{O}(\frac{1}{\epsilon^3})$, $\beta=\mathcal{O}(\epsilon^2)$, and $d_i=\mathcal{O}(\frac{1}{\epsilon})$ yields an $\mathcal{\epsilon}$-FOSP for the MAML-based FL.
\end{IEEEproof}

\begin{rem} \label{rem:1}
It is worth noting that the result in Corollary~\ref{cor:1} provides a possible choice of the hyperparameters. Other options may also be valid as long as the option makes the order of magnitude of the hyperparamters (w.r.t. the model accuracy $\epsilon$) satisfy the condition in Theorem \ref{thm:1}.
\end{rem}

Guided by Corollary \ref{cor:1}, we can recast (C1.1) in problem (P1) as a series of constraints, which give explicit relationships between optimization variables and the model accuracy $\epsilon$. In this way, we then solve problem (P1) with high efficiency in the next section.

\section{Automated Federated Learning (AutoFL)}\label{sec:4}

In this section, we first resort to Theorem~\ref{thm:1} and Corollary~\ref{cor:1} for problem (P1) decomposition, and then solve the resultant subproblems respectively. Based on the obtained solutions, we finally propose Automated Federated Learning (AutoFL) algorithm, by which the benefit MAML brings to FL can be quantified in mobile edge networks.

\subsection{Problem Decoupling} \label{sec:4.1}

The clue to the decomposition of problem (P1) follows the primal decomposition~\cite{palomar2006tutorial}. Specifically, the master problem is optimized with respect to the number of communication rounds $K$, which is given as follows:

\begin{subequations}\label{equ:opt2}
  \begin{align}
    \min_K \qquad & \sum_{k=0}^{K-1} \max_{i\in\mathcal{U}}\{t_{k,i}^{\text{cmp}} + t_{k,i}^{\text{com}}\} \label{equ:opt2_obj}\tag{P2}\\
    \text{s.t.} \qquad & \frac{4(F(w_0) - F(w_{\epsilon}))}{\beta K} \leq \epsilon, \qquad K\in \mathbb{N}^+. \label{equ:opt2.1}\tag{C2.1}
  \end{align}
\end{subequations}
Note that, we refer to Theorem~\ref{thm:1} and use a contracted (C1.1), i.e., (C2.1), as the constraint in problem (P2) for tractability. This contraction replaces $\frac{1}{K} \sum_{k=0}^{K-1} \mathbb{E}[\|\nabla F(\bar{w}_k)\|^2]$ with its upper bound without the second term. The reason that we get rid of the second term is that we only consider $K$ as the decision variable in (P2), and the second term of the upper bound can be regarded as a constant and thus removed as long as its value is no larger than $\epsilon$.
Meanwhile, the above contraction shrinks the feasible region of $K$. Therefore, as long as (\ref{equ:opt2.1}) is satisfied, (\ref{equ:opt1.1}) must be satisfied. Moreover, if there exists an optimal solution $K$ to problem (P2), it is also feasible to the original problem (P1). We can regard the optimal solution to (P2) as an optimal approximation to (P1).

The reason that we consider $K$ as a decision variable is for the decoupling of (P1). The original objective of minimizing the overall learning time in $K$ rounds can be regarded as the minimization of the learning time in each round, once $K$ is determined and is set to be a constant in the following decoupling. As such, problem (P2), which has a slave problem with respect to the sampled data size $\mathbf{d}$ and transmit power $\mathbf{p}$, can be regarded as an approximation of problem (P1).

From Theorem \ref{thm:1}, we find that $\mathbf{d}$ and $\mathbf{p}$ are coupled even though (C1.1) is replaced by the right-hand-side term in (\ref{equ:thm1}). In this regard, we aim to optimize $\mathbf{d}$ and $\mathbf{p}$ in the slave problem with the \textit{coordinate descent}~\cite{wright2015coordinate} method. That is, the slave problem is firstly solved with $\mathbf{d}$ as variables and $\mathbf{p}$ as constants, and then is solved with $\mathbf{p}$ as variables and $\mathbf{d}$ as constants. This procedure is repeated in an iterative manner. In this way, $\mathbf{d}$- and $\mathbf{p}$-related slave problems can be further decomposed among rounds under the assistance of Corollary \ref{cor:1}. Specifically, in any round $k$, the sampled data size for UE $i$ is optimized by solving problem (P3):


\begin{subequations}\label{equ:opt3}
  \begin{align}
    \min_{d_k^i} \qquad & \max_i \quad \left\{\frac{c_i}{\vartheta_i} d_k^i  + t_{k, i}^{\text{com}} \right\} \label{equ:opt3_obj}\tag{P3}\\
    \text{s.t.} \qquad & d_k^i \geq \frac{1}{\epsilon}, \quad \forall i\in \mathcal{U}, \label{equ:opt3.1}\tag{C3.1} \\
    &  \frac{\varsigma}{2}c_i d_k^i\vartheta_i^2 + p_k^{i}\frac{Z}{B \log_2 (1 + \xi_k^i)}\leq E_{\max}, \quad \forall i\in \mathcal{U}, \label{equ:opt3.2}\tag{C3.2} \\
    & 0 \leq d_k^i \leq D_i, \quad \forall i \in \mathcal{U}. \label{equ:opt3.3}\tag{C3.3}
  \end{align}
\end{subequations}
(\ref{equ:opt3.1}) is obtained from Corollary~\ref{cor:1}, that a $d_i=\mathcal{O}(\frac{1}{\epsilon})$ will yield an $\epsilon$-FOSP. (\ref{equ:opt3.1}) uses $d_k^i \geq \frac{1}{\epsilon}$ instead to indicate that as long as $d_k^i$ is not smaller than $\frac{1}{\epsilon}$, an $\epsilon$-FOSP would be guaranteed.
This is a reasonable replacement (i.e., from $d_k^i = \mathcal{O}(\frac{1}{\epsilon})$ to $d_k^i \geq \frac{1}{\epsilon}$) such that, the larger the sampled data size is, the larger accuracy AutoFL would achieve, thereby leading to a smaller $\epsilon$.
Note that (\ref{equ:opt3.1}) is also a contraction of (\ref{equ:opt1.1}), and the feasible domain of $d_k^i$ is also shrunk. That is, the optimal $d_k^{i*}$, as long as it exists in (P3), is also feasible in problem (P1), and can be regarded as an optimal approximation to (P1).
(\ref{equ:opt3.2}) is the energy consumption constraint on each UE and (\ref{equ:opt3.3}) indicates that the sampled data size of UE $i$ is no greater than the number of data points generated by UE $i$.
Furthermore, addressing the following problem can make an decision on the transmit power of UE $i$ in round $k$:
\begin{subequations}\label{equ:opt4}
  \begin{align}
    \min_{p_k^i} \qquad & \max_{i} \quad \left\{t_{k,i}^{\text{cmp}}+ \frac{Z}{B \log_2 (1+ \xi_k^i)}\right\}
    \label{equ:opt4_obj}\tag{P4} \\
    \text{s.t.} \qquad & \left(\sum_{i=1}^{n} {U_k^i}^2\right) \frac{\alpha^2 L^2 \sigma_G^2}{D^{\text{in}}} \leq \epsilon, \label{equ:opt4.1}\tag{C4.1} \\
    & \frac{\varsigma}{2}c_id_k^i\vartheta_i^2 + p_k^i\frac{Z}{B \log_2(1 + \xi_k^i)} \leq E_{\max}, \quad \forall i\in \mathcal{U}, \label{equ:opt4.2}\tag{C4.2} \\
    & 0 \leq p_k^i \leq P_{\max}, \quad \forall i\in \mathcal{U}, \label{equ:opt4.3}\tag{C4.3}
  \end{align}
\end{subequations}
where (\ref{equ:opt4.1}) is obtained from Corollary~\ref{cor:1}, in terms of the fact that $(\sum_{i=1}^{n} {U_k^i}^2) \frac{\alpha^2 L^2 \sigma_G^2}{D^{\text{in}}}$ dominates the value of the second term in the right-hand-side of (\ref{equ:thm1}) to guarantee an $\epsilon$-FOSP for the MAML-based FL. Here we replace $U_i$ with $U_k^i$ to show that in each round $k$, (\ref{equ:opt4.1}) should be satisfied.
The reason that we replace $(\sum_{i=1}^{n} {U_k^i}^2) \frac{\alpha^2 L^2 \sigma_G^2}{D^{\text{in}}} = \mathcal{O}(\frac{1}{\epsilon})$ with $(\sum_{i=1}^{n} {U_k^i}^2) \frac{\alpha^2 L^2 \sigma_G^2}{D^{\text{in}}} \leq \epsilon$ is to indicate that, as long as (C4.1) is satisfied, an $\epsilon$-FOSP would be guaranteed.
Note that (\ref{equ:opt4.1}) is also a contraction of (\ref{equ:opt1.1}). That is, as long as there exists an optimal solution to (P4), this solution is also feasible and is close to the optimal solution to (P1). We can regard the optimal solution to (P4) as an optimal approximation to (P1).
(\ref{equ:opt4.2}) and (\ref{equ:opt4.3}) are the energy constraint and transmit power constraint on each UE, respectively.

Based on the above decomposition, we can solve the original problem (P1) by solving problem (P2), in which problems (P3) and (P4) are nested and addressed with the coordinate descent method. This decoupling only provides the approximate optimal results. However, the evaluation results show that, with the approximate optimal results obtained from (P2), (P3) and (P4), the proposed MAML-based FL algorithm always outperforms Per-FedAvg in learning time and convergence performances. Next, we deal with problems (P2), (P3), and (P4), respectively.

\subsection{Problem Solution}
\subsubsection{Communication Round Optimization} \label{sec:4.2}

Once the power transmit power $\mathbf{p}$ and the sampled data size $\mathbf{d}$ are obtained from the slave problem, the master problem (P2) is monotonously increasing with respect to $K$. Therefore, to minimize the total training time, the optimal value of $K$ should be obtained as its lower bound:
\begin{equation}\label{equ:optimal_K}
  K^* = \frac{4(F(w_0) - F(w_{\epsilon}))}{\beta \epsilon},
\end{equation}

Note that, we do not use the formulation of $K^*$ to predict the actual optimal global communication rounds. It is not practical to determine this value in advance because in practice, the optimal value of the communication rounds can be easily observed once the training loss starts to converge. There are so many factors that can affect the actual value of $K^*$, even the version of Pytorch/Tensorflow in the experiments! Therefore, the theoretical formulation of $K^*$ in (31), with the initial global model $w_0$, the global optimal loss $F(w_{\epsilon})$, the global learning rate $\beta$, and the model accuracy $\epsilon$ to be decided, is only used as a guidance indicating the order of magnitude of the practical optimal communication rounds.

\subsubsection{Data Sample Size Optimization} \label{sec:4.3}

It is easy to observe from problem (P3) that for each UE $i$, its optimal $d_k^i$ lies at the lower bound $\frac{1}{\epsilon}$.
However, whether this lower bound can be reached depends on the relationship between the values of $d_k^i$'s lower bound $\frac{1}{\epsilon}$ and upper bounds defined by (\ref{equ:opt3.2}) and (\ref{equ:opt3.3}). Specifically, the upper bound defined by (\ref{equ:opt3.2}) is $D_i^{\text{p}}:= \frac{2E_{\max}}{\varsigma c_i\vartheta_i^2} - \frac{2p_k^i Z}{\varsigma c_i\vartheta_i^2 B \log_2(1 + \xi_k^i)}$, while the upper bound defined by (\ref{equ:opt3.3}) is $D_i$.
%
%
Consequently, for each UE $i$, we need to consider two cases: $D_i \geq \frac{1}{\epsilon}$ and $D_i < \frac{1}{\epsilon}$, in which we further discuss other two cases: $D_i^{\text{p}}>\frac{1}{\epsilon}$ and $D_i^{\text{p}}\leq\frac{1}{\epsilon}$.
That is, we discuss the optimal solution of $d_k^i$ for problem (P3) as follows.

\vspace{0.4cm}
\noindent\textbf{Case 1}: In this case, UE $i$ generates sufficient data points which are larger than or equal to its lower bound for a model accuracy $\epsilon$. That is, for each UE $i$, we have
\begin{equation}\label{equ:extreme1}
  0 < \frac{1}{\epsilon}\leq D_i.
\end{equation}
Under this case, we need to further discuss the relationship between $D^{\text{p}}_i$ and $\frac{1}{\epsilon}$:

\begin{itemize}
  \item When $D_i^{\text{p}} > \frac{1}{\epsilon}$, the lower bound $\frac{1}{\epsilon}$ of $d_k^i$ can be obtained with enough data points to achieve the model accuracy $\epsilon$. Therefore, the optimal sampled data size $d_k^i = \frac{1}{\epsilon}$.
  \item When $D_i^{\text{p}} \leq \frac{1}{\epsilon}$, the lower bound $\frac{1}{\epsilon}$ of $d_k^i$ cannot be achieved. According to (\ref{equ:thm1}) in Theorem \ref{thm:1}, the larger $d_i$ is, the higher model accuracy can be achieved. Therefore, the optimal value of $d_k^i$ should be equal to $D_i^{\text{p}}$.
\end{itemize}

\vspace{0.4cm}
\noindent\textbf{Case 2}: In this case, UE $i$ creates deficient data points which are smaller than its lower bound for a model accuracy $\epsilon$. That is, for each UE $i$, we have
\begin{equation}\label{equ:extreme2}
  0 \leq D_i < \frac{1}{\epsilon}.
\end{equation}
Under this case, we also need to further discuss the relationship between $D^{\text{p}}_i$ and $\frac{1}{\epsilon}$:
\begin{itemize}
  \item When $D^{\text{p}}_i > \frac{1}{\epsilon}$, the optimal value of $d_k^i$ is $D_i$;
  \item When $D^{\text{p}}_i \leq \frac{1}{\epsilon}$, the optimal value of $d_k^i$ is $\min\{D_i,D_i^{\text{p}}\}$;
\end{itemize}

The above process of computing $d_k^i$ for problem (P3) can be summarized as \texttt{SampledDataSize} algorithm, which is omitted due to the space limitation.

\subsubsection{Transmit Power Optimization} \label{sec:4.4}
At last, we turn to solving problem (\ref{equ:opt4_obj}), where transmit power $p_k^i$ for UE $i$ in round $k$ is optimized with fixed $d_k^i$. To this end, we first convert the problem into a more concise form. Given that $U_k^i$ denotes the normalized update successful probability, we have $U_k^i \leq 1$. Then the inequality $\sum_i^n {U_k^i}^2 \leq 1$ is always true. Consequently, we can further transform (\ref{equ:opt4.1}) to $\frac{\alpha^2 L^2 \sigma_G^2}{D_i^{\text{in}}} \leq \epsilon$. Combine the transformed constraint with (\ref{equ:opt3.1}), (\ref{equ:opt3.2}) and (\ref{equ:opt3.3}), we find that as long as $\alpha^2 L^2 \sigma_G^2 \leq \min\{1, D_i\epsilon, D_p\epsilon\}$, (\ref{equ:opt4.1}) can always be satisfied. Therefore, problem (\ref{equ:opt4_obj}) can be further transformed to a new optimization problem without constraint (\ref{equ:opt4.1}).

With the above concise formulation of (\ref{equ:opt4_obj}), we analyze the monotonicity of $t_{k,i}^{\text{com}}$ in the objective function of problem (\ref{equ:opt4_obj}). Specifically, the derivative of $t_{k,i}^{\text{com}}$ can be computed as follows:
\begin{align}\label{equ:derivative_p3}
    \frac{\text{d}}{\text{d} p_k^i} t_{k,i}^{\text{com}} 
    = & \frac{\text{d}}{\text{d} p_k^i} \frac{Z}{B \log_2 (1+ \xi_k^i)} \nonumber \\
    = & -\frac{Z h_i \|c_i\|^{-\kappa}}{B N_0(1+\xi_k^i)[\log_2 (1+ \xi_k^i)]^2}<0,
\end{align}
which means $t_{k,i}^{\text{com}}$ monotonically decreases with $p_k^i$.

Further, as for the constraint (C4.2), the derivative of its left-hand-side term can be proved to be always larger than 0, which is shown as follows,
\begin{align}
   & \frac{\text{d}}{\text{d} p_k^i} p_k^i\frac{Z}{B \log_2(1 + \xi_k^i)} \nonumber \\
   = & \frac{Z}{B} \left( \frac{1}{\log_2(1 + \xi_k^i)} - \frac{\xi_k^i}{[\log_2(1 + \xi_k^i)]^2(1 + \xi_k^i)}\right) \nonumber \\
  > &  \frac{Z}{B\log_2(1 + \xi_k^i)} \left( 1 - \frac{\xi_k^i}{\frac{\xi_k^i}{1 + \xi_k^i}(1 + \xi_k^i)} \right) \nonumber \\
  = & \frac{Z}{B\log_2(1 + \xi_k^i)} (1-1)=  0
\end{align}
where the inequality is derived from the fact that $\log_2(1+x) > \frac{x}{1+x}$, for $x > 0$. Therefore, the left-hand-side term of (C4.2) monotonically increases with $p_k^i$. It means (C4.2) defines another upper bound of $p_k^i$ just as $P_{\max}$ in (C4.3).

From the above analysis, the optimal solutions to (\ref{equ:opt4_obj}) lies at the upper bound of $p_k^i$, which is defined by (\ref{equ:opt4.2}) or (\ref{equ:opt4.3}). This process of computing $p_k^i$ can be summarized as \texttt{PowerComputation} algorithm, which is omitted due to space concern.

\subsection{AutoFL Implementation} \label{sec:4.5}

\begin{algorithm}[h]
    \caption{Automated Federated Learning (AutoFL)}
    \label{alg:AutoFL}
    \SetKwInOut{Input}{Input}
    \Input{\begin{minipage}[t]{0.9\linewidth}
        $\phi$, $E_{\max}$, $P_{\max}$, $N_0$, $c_i$, $B^{\text{up}}$, $\kappa$, $\vartheta_i$, $w_0$.
        \end{minipage}}
    $p_{-1}^i := P_{\max}$;\\
    \For{$k=0,1,\dots,K-1$}{
        The BS randomly selects a subset of associated UEs $\mathcal{A}_k$ where each UE $i\in\mathcal{A}_k$ is with $s_{k}^i = 1$\;
        The BS sends $w_k$ to all UEs in $\mathcal{A}_k$ \;
        \For{$i\in\mathcal{A}_k$}{
            $d_k^i := \texttt{SampledDataSize}(p_{k-1}^i)$ \;
            $p_k^i := \texttt{PowerComputation}(d_k^i)$ \;
            \If{$\gamma_k^i \geq \phi$}{
              $w_{k+1}^i := \texttt{LocalModelTraining}(d_k^i, w_k)$ \;
              UE $i$ sends $w_{k+1}^i$ back to the BS \;
            }

        }
        The BS updates its model using $w_{k+1} := \frac{1}{\sum_{i=1}^{n} \mathds{1}\{s_{k}^i = 1, \gamma_k^i>\phi\}} \sum_{i=1}^{n} \mathds{1}\{s_{k}^i = 1, \gamma_k^i\geq\phi\} w_{k+1}^i$ \;
    }
    If the model accuracy $\epsilon$ is reached, terminate the algorithm and set $K^* := K$
\end{algorithm}

After devising the algorithms to determining the number of communication round $K$, the sampled data size $\mathbf{d}$, and the transmit power $\mathbf{p}$, we are able to combine these sub-algorithms together to design the AutoFL algorithm, which is shown in Algorithm~\ref{alg:AutoFL}. Note that, to reduce the computational complexity, we only adopt the coordinate descent method to solve problem (P3) and (P4) once in one round. More specifically, given the current transmit power $p_{k-1}^i$, problem (P4) with respect to $d_{k}^i$ is solved for UE $i\in\mathcal{A}_k$ in round $k$ using \texttt{SampledDataSize} algorithm, as line 6 in AutoFL. Then, based on the obtained $d_k^i$, problem (P4) with respect to $p_k^i$ is addressed for UE $i\in\mathcal{A}_k$ in round $k$ with \texttt{PowerComputation} algorithm, as line 7 in AutoFL. The result of $p_k^i$ is further used in the next round $k+1$ to compute $d_{k+1}^i$ for UE $i\in\mathcal{A}_k$, which is used to compute $p_{k+1}^i$ again. This process repeats until model accuracy $\epsilon$ is achieved. As in line 15 in AutoFL, once the model accuracy is reached, the algorithm is terminated and the required number of rounds $K^{*}$ is output.

\section{Performance Evaluation} \label{sec:5}

In this section, we evaluate the performances of AutoFL to
(1) verify its effectiveness in saving the learning time, (2) examine its training loss and accuracy to demonstrate its effectiveness in fast adaptation and convergence, (3) present its advantages in different network and dataset settings.

\subsection{Simulation Settings} \label{sec:5.1}

\subsubsection{Network settings}
\begin{table*}
\centering
\caption{System Parameters}
\label{tab:sys_para}
\begin{tabular}{c|c||c|c||c|c}
  \hline
  \textbf{Parameter} & \textbf{Value} & \textbf{Parameter} & \textbf{Value} & \textbf{Parameter} & \textbf{Value}\\
  \hline
  $\alpha$ (MNIST) & $0.03$ & $B$ & 1 MHz & $E_{\max}$ & $0.003$ J\\
  \hline
  $\beta$ (MNIST) & $0.07$ & $\varsigma$ & $10^{-27}$ & $P_{\max}$ & $0.01$ W \\
  \hline
  $\alpha$ (CIFAR-10) & 0.06 & $\kappa$ & $3.8$ & $\phi$ & $30$ dB\\
  \hline
  $\beta$ (CIFAR-10) & 0.02 & $\vartheta_i$ & $10^9$ & $N_0$ & $-174$ dBm/Hz\\
  \hline
\end{tabular}
\end{table*}

Unless otherwise sepecified, we consider a mobile edge network that consists of $n=20$ UEs located in a cell of radius $R = 200$ m and a BS located at its center.
Assume that all UEs are uniformly distributed in the cell.
The Rayleigh distribution parameter of $h_k^i$ is $40$.
The other parameters used in simulations are listed in Table~\ref{tab:sys_para}.

\subsubsection{Datasets}
We evaluate the training performance using two datasets, the MNIST dataset~\cite{MNIST} for handwritten digits classification and CIFAR-10~\cite{CIFAR10}. The network model we used for MNIST is a 2-layer NN with hidden layer of size 100, while the network model we used for CIFAR-10 is LeNet-5~\cite{lecun1998gradient}, where there are two convolutional layers and three fully connected layers. The datasets are partitioned randomly with $75\%$ and $25\%$ for training and testing respectively. Meanwhile, in the simulation we use both i.i.d datasets and non-i.i.d datasets.
In order to generate i.i.d datasets, the original training dataset is randomly partitioned into 20 portions and each UE is assigned a portion. As for the non-i.i.d datasets, the original training set is first partitioned into 10 portions according to the labels. Each UE has only $l$ of the 10 labels and is allocated a different local data size in the range of $[2, 3834]$, where $l = [1,\dots,10]$. The value of $l$ reflects the non-i.i.d level of local datasets. The smaller $l$ is, the higher non-i.i.d level.
Unless otherwise defined, we use $l=5$ in the evaluation.
%
All experiments are conducted by PyTorch~\cite{paszke2019pytorch} version 1.7.1.

\begin{figure*}[t]
  \centering
  \subfloat[I.i.d MNIST datasets]{
      \includegraphics[width=2.3in]{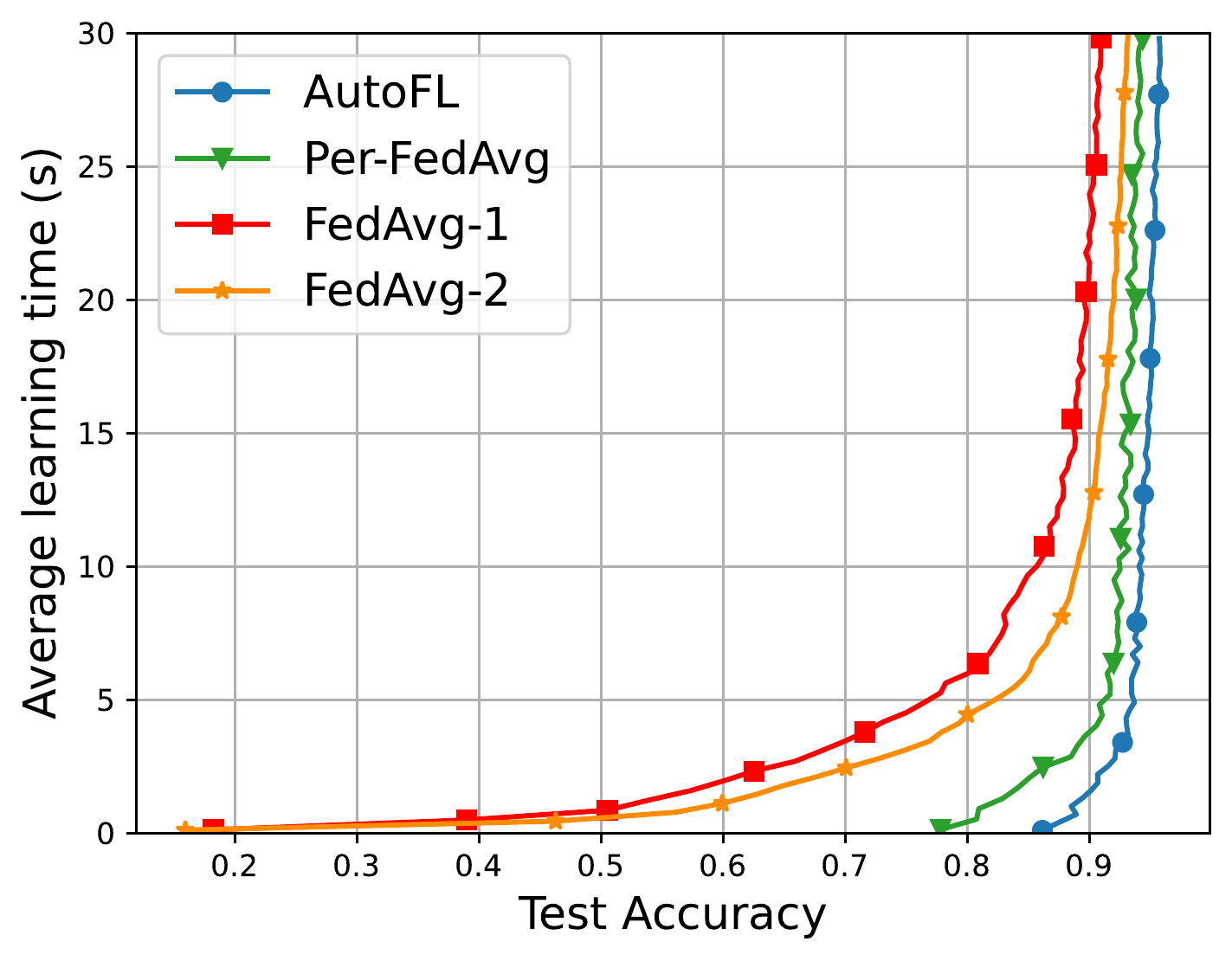}
      \label{fig:runtime:subfig:a}}
  \subfloat[Non-i.i.d MNIST datasets]{
      \includegraphics[width=2.3in]{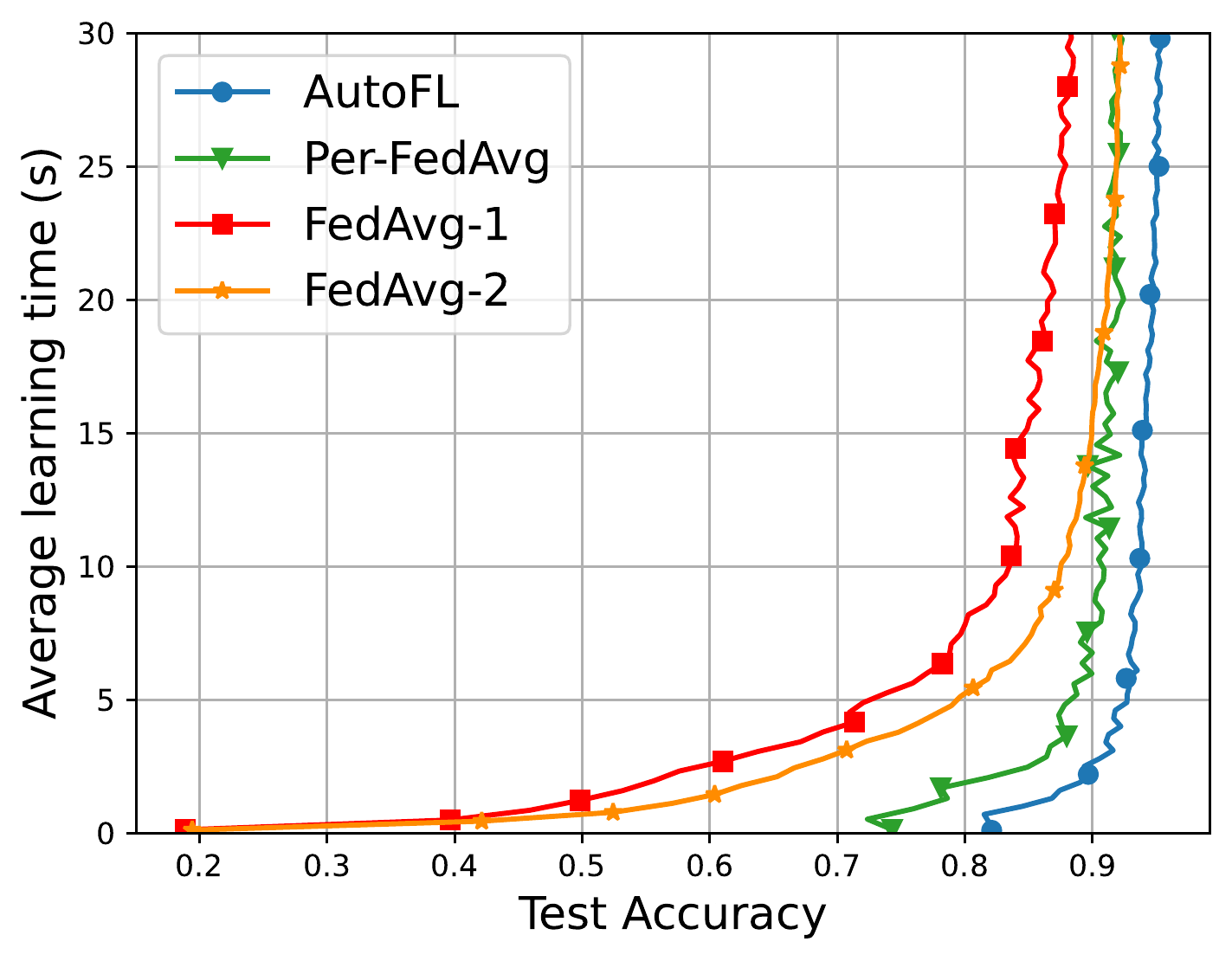}
      \label{fig:runtime:subfig:b}}
  \subfloat[Non-i.i.d CIFAR-10 datasets]{
      \includegraphics[width=2.3in]{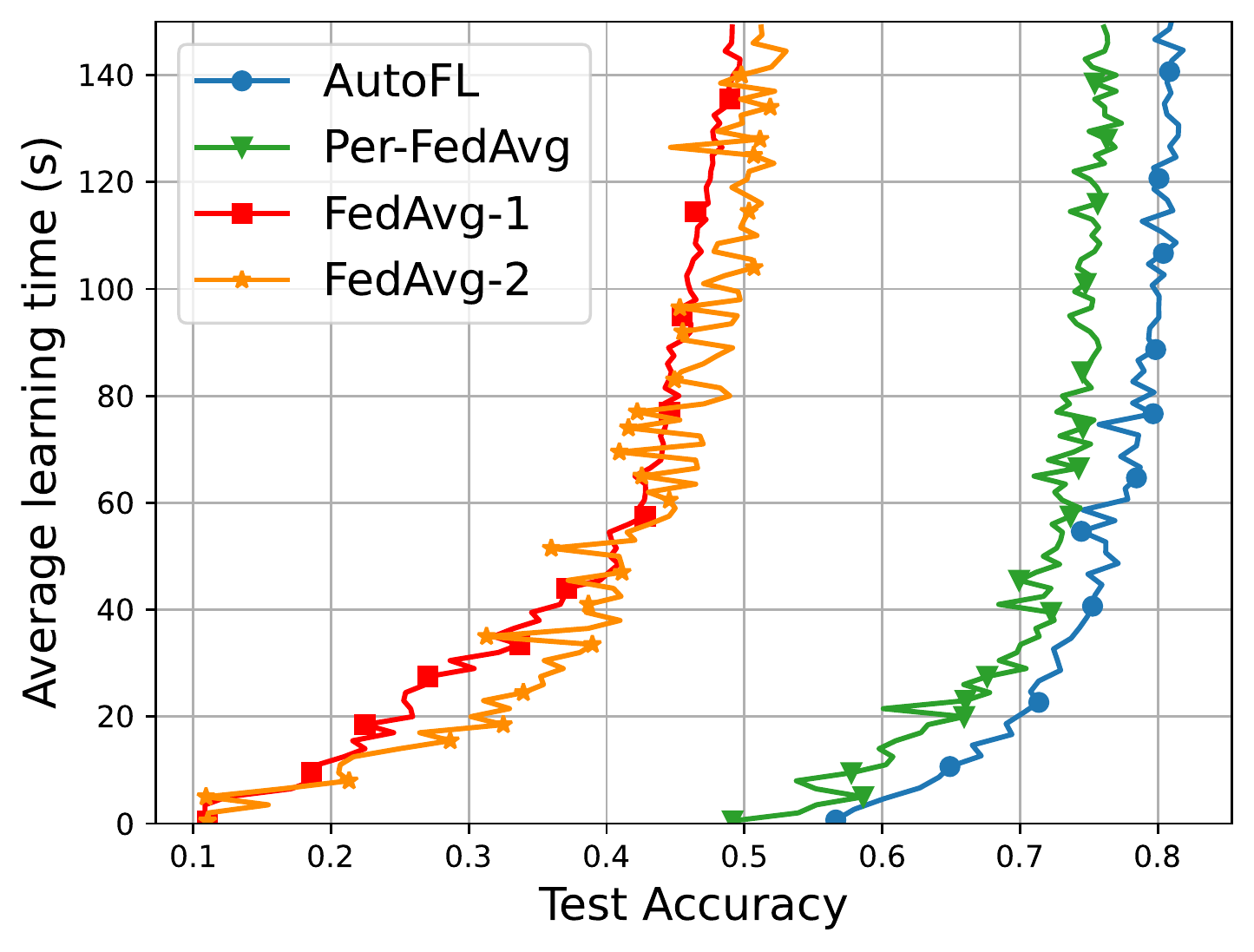}
      \label{fig:runtime:subfig:C}}
  \caption{Learning time comparisons for i.i.d MNIST, non-i.i.d MNIST and non-i.i.d CIFAR-10 datasets. As for the non-i.i.d MNIST and CIFAR-10 datasets, $l=5$.}
  \label{fig:runtime}
\end{figure*}

\begin{figure}[t]
  \centering
  \subfloat[Non-i.i.d MNIST dataset]{
      \includegraphics[width=2.3in]{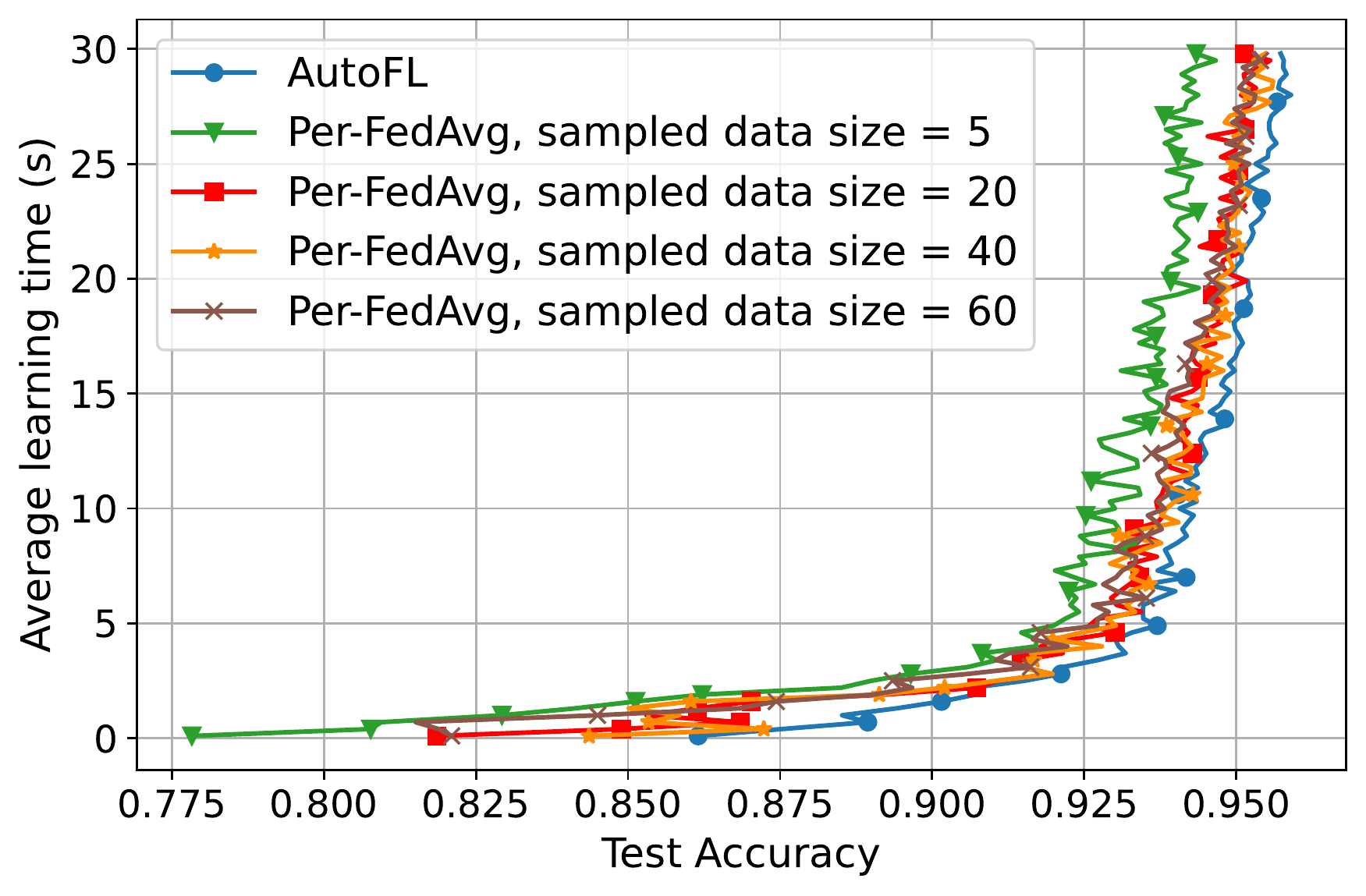}
      \label{fig:sampled_data_size:subfig:a}}
  \hfil
  \subfloat[Non-i.i.d CIFAR-10 dataset]{
      \includegraphics[width=2.3in]{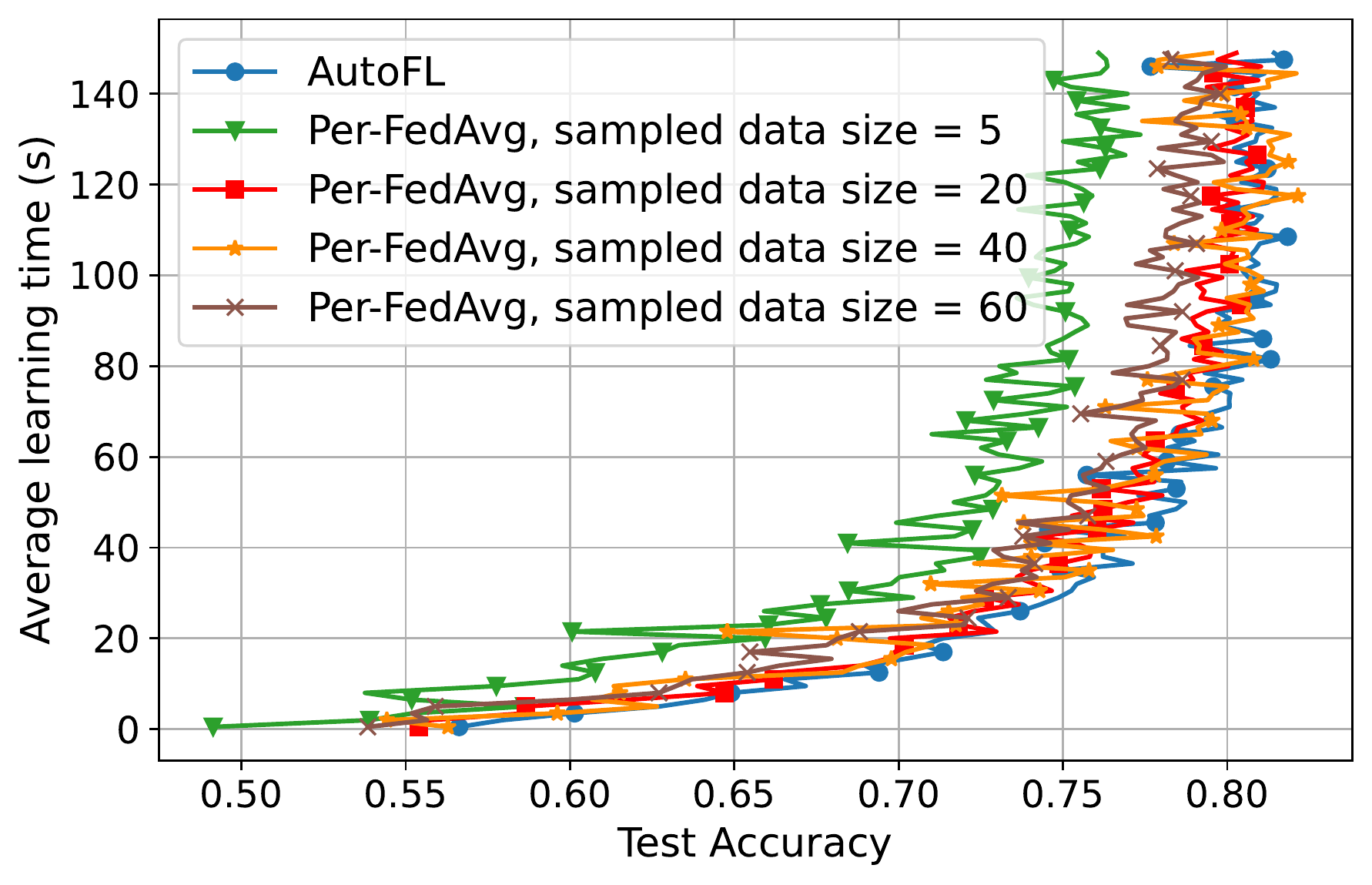}
      \label{fig:sampled_data_size:subfig:b}}
  \caption{Learning time comparisons between AutoFL and Per-FedAvg with respect to different sampled data sizes using non-i.i.d MNIST and CIFAR-10 datasets.}
  \label{fig:sampled_data_size}
\end{figure}

\subsubsection{Baselines}
We compare the performance of AutoFL with Per-FedAvg~\cite{fallah2020personalized} and FedAvg~\cite{mcmahan2017communication}.
FedAvg is the first algorithm that is proposed for FL and is the most general FL algorithm. Per-FedAvg, on the other hand, is the most general MAML-based FL algorithm.
In Per-FedAvg, we first fix the sampled data size of all UEs to be $5$.
As for the transmit power settings in Per-FedAvg, all UEs use the maximal power $P_{\max}$, given that there are only 5 data samples to train on each UE and more power can be saved to transmit the local updates to the BS.
As for FedAvg, we consider two sets of sampled data sizes: for the first one, we set the sampled data sizes of the UEs to be $5$ and the transmit power in all rounds to be $P_{\max}$, which is the same with the settings with Per-FedAvg.
We name the FedAvg algorithm with the first setting as FedAvg-1;
For the second one, we set the sampled data sizes and transmit power of the UEs to be the same with AutoFL.
We name the FedAvg with the second setting as FedAvg-2.
Later, we may change the sampled data sizes in Per-FedAvg and FedAvg-1 to check the influence of the number of sampled data on the performance gap between Per-FedAvg and AutoFL.

\subsection{Learning Time Comparisons} \label{sec:5.2}

We first compare the learning time of AutoFL with that of Per-FedAvg, FedAvg-1, and FedAvg-2 with respect to the test accuracy.
The results are shown in Fig.~\ref{fig:runtime}. We analyse the results from three aspects. (i) First, it is clear that at the same accuracy, AutoFL consumes the smallest learning time.
For example, when the algorithms start to converge, Per-FedAvg takes at least twice as much time as AutoFL.
This is a reasonable result as AutoFL is designed to minimize the overall learning time.
Meanwhile, as the test accuracy increases, the average learning time of all algorithms grows rapidly.
The experimental results verify our theoretical analysis and confirms the effectiveness of AutoFL in saving the overall learning time. (ii) Second, we observe that the two MAML-based FL algorithms outperform the two conventional FL algorithms. This advantage is even more pronounced when using the non-i.i.d CIFAR-10 dataset rather than the MNIST datasets. This result testifies the advantage of MAML in accelerating the training speed of FL.
Meanwhile, the two algorithms with designed sampled data size and transmit power, AutoFL and FedAvg-2 outperform Per-FedAvg and FedAvg-1. This indicates that the joint optimization of the sampled data size and transmit power is a promising way to improve the FL performance over mobile edge networks. (iii) Third, the results derived from i.i.d datasets are more stable than that from non-i.i.d datasets, meanwhile the average learning time under i.i.d datasets is smaller than that under non-i.i.d datasets. This is because, in the i.i.d case, the local model of each individual UE has a higher representative for the learned global model, which is beneficial to improve the learning performance.

\begin{figure*}[t]
  \centering
  \subfloat[I.i.d MNIST dataset]{
      \includegraphics[width=2.3in]{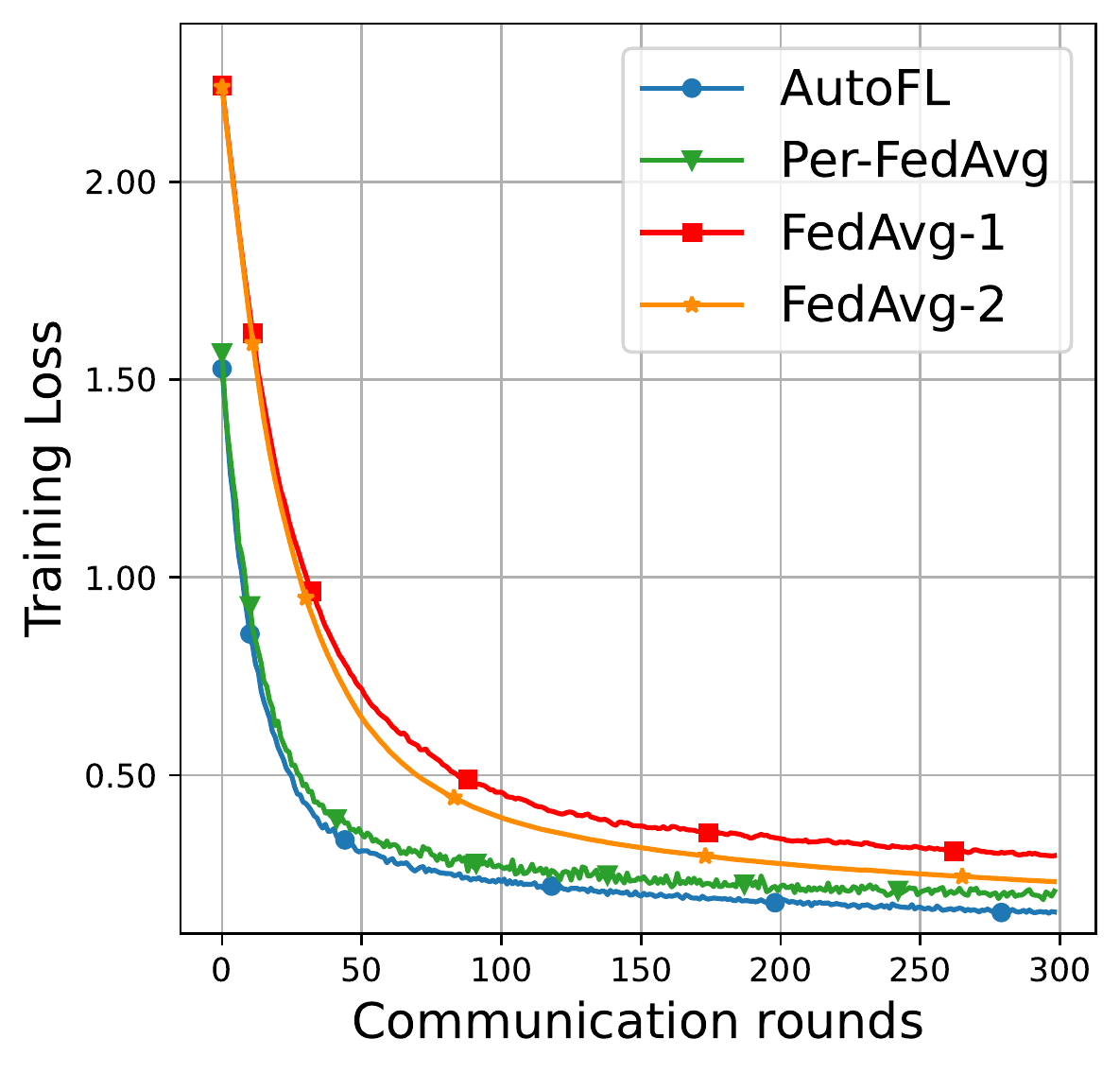}
      \label{fig:convergence:subfig:a}}
  \subfloat[Non-i.i.d MNIST dataset]{
      \includegraphics[width=2.3in]{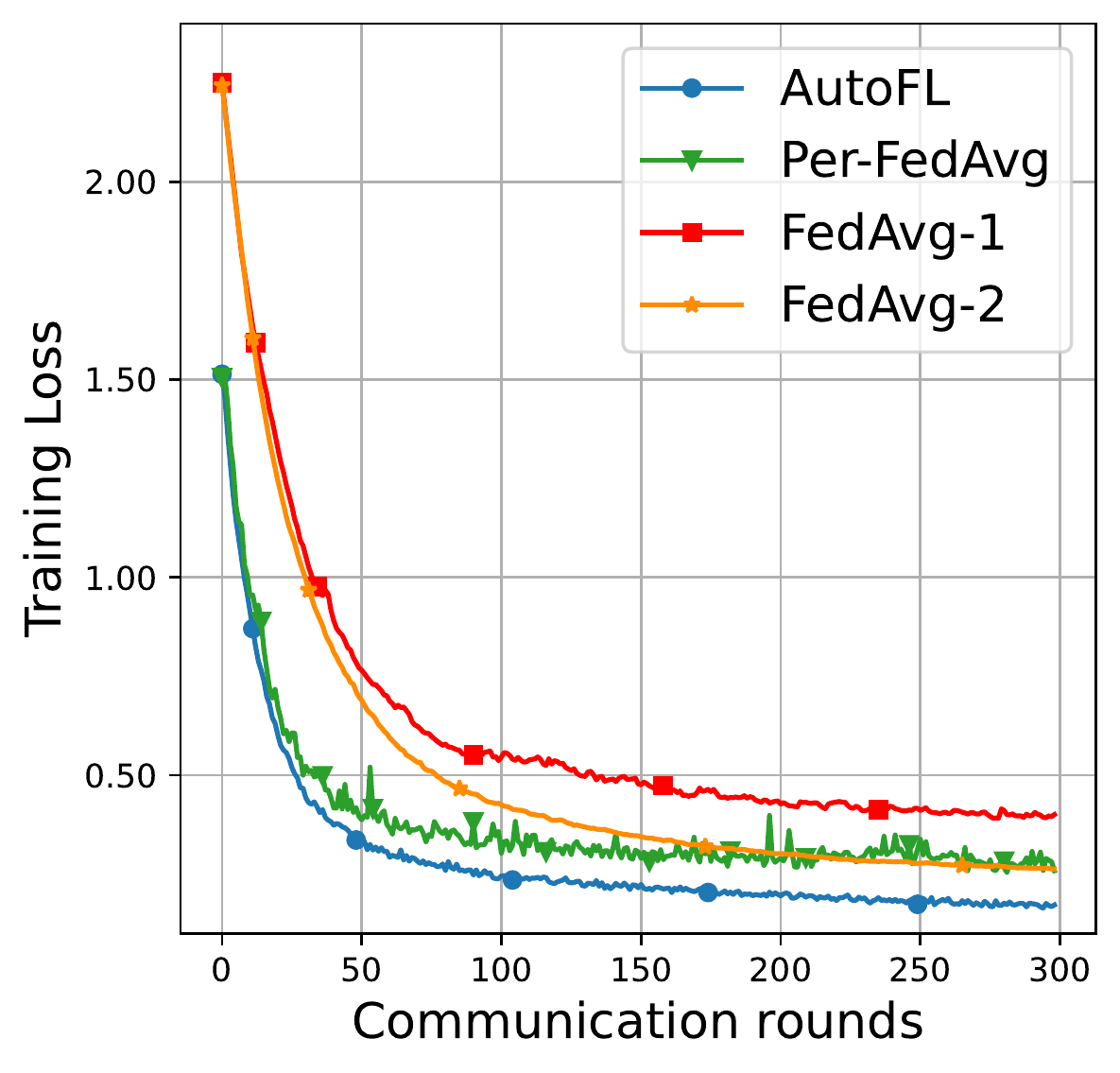}
      \label{fig:convergence:subfig:b}}
  \subfloat[Non-i.i.d CIFAR-10 dataset]{
  \includegraphics[width=2.3in]{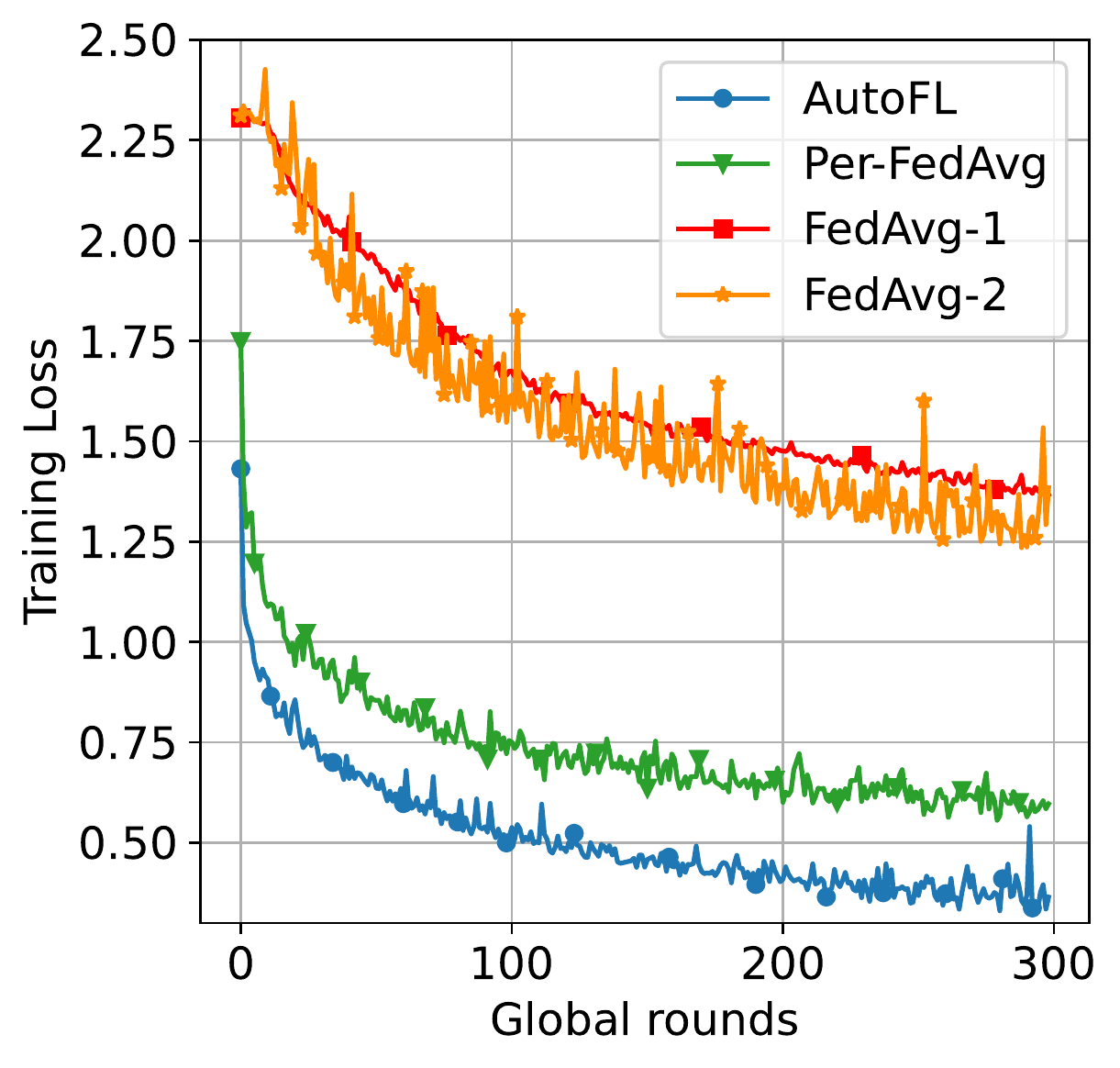}
      \label{fig:convergence:subfig:c}}
  \hfil
  \subfloat[I.i.d MNIST dataset]{
      \includegraphics[width=2.3in]{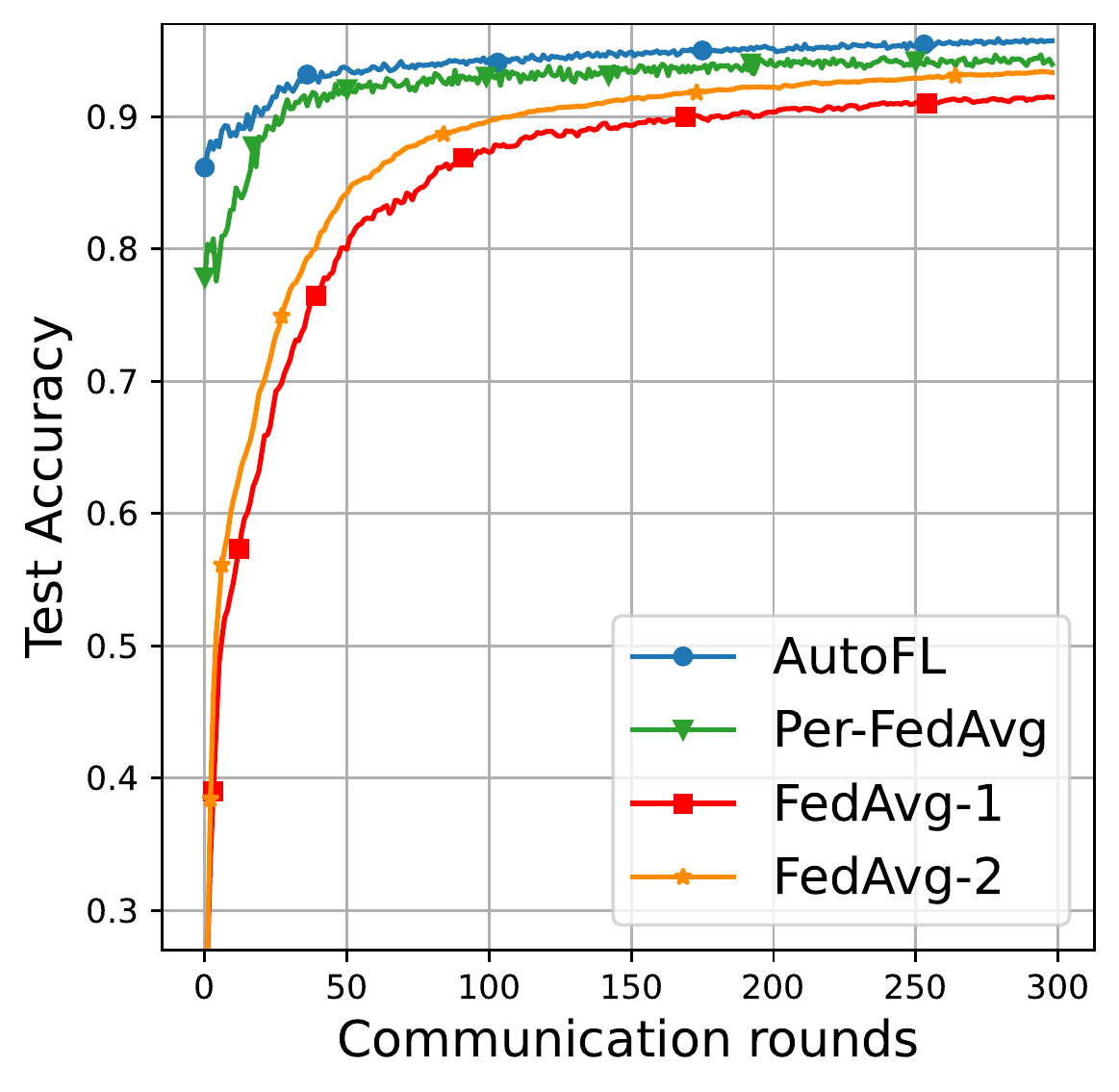}
      \label{fig:convergence:subfig:d}}
  \subfloat[Non-i.i.d MNIST dataset]{
      \includegraphics[width=2.3in]{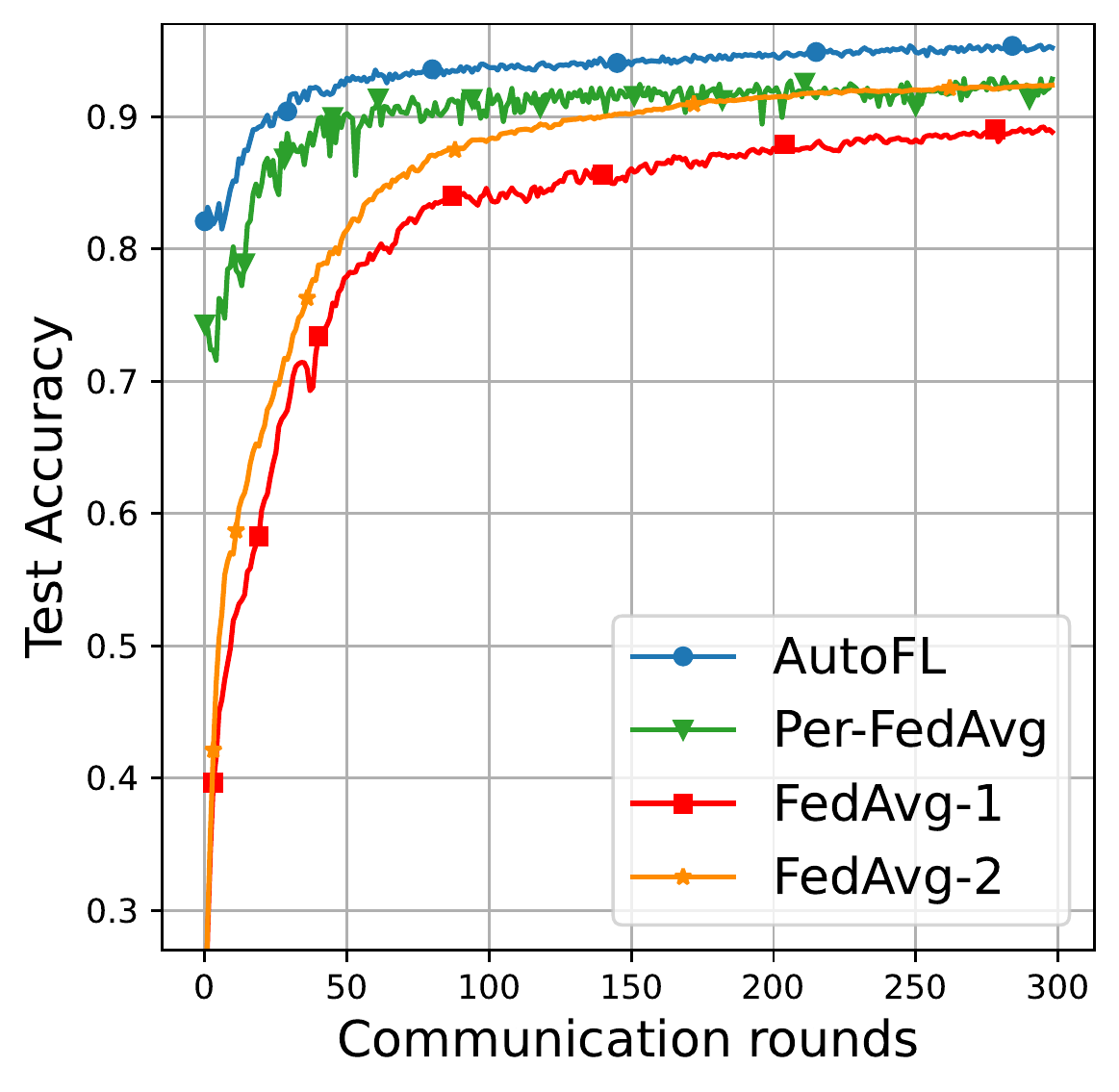}
      \label{fig:convergence:subfig:e}}
  \subfloat[Non-i.i.d CIFAR-10 dataset]{
  \includegraphics[width=2.3in]{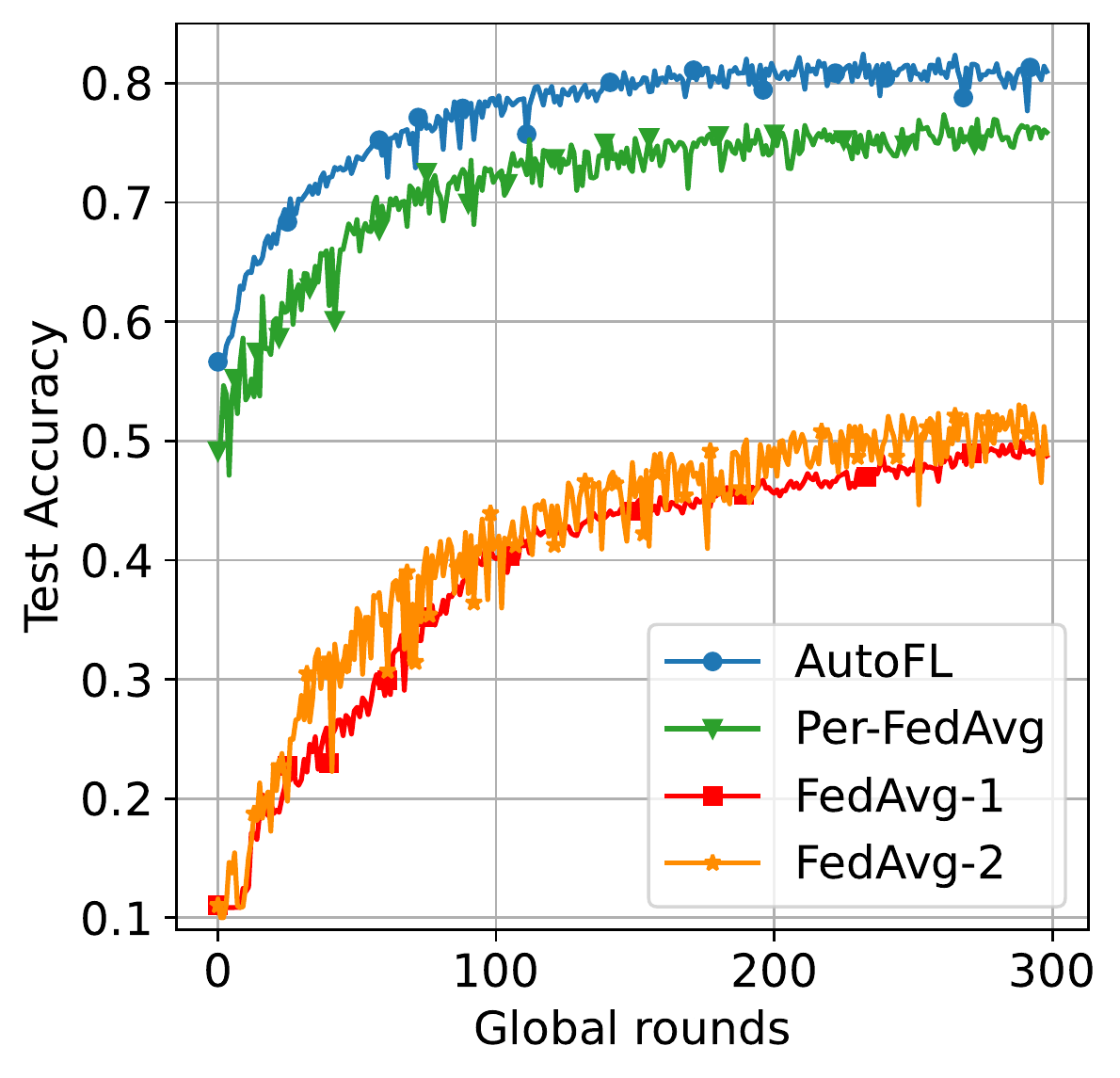}
      \label{fig:convergence:subfig:f}}
  \caption{Convergence performance comparisons using i.i.d, non-i.i.d MNIST and non-i.i.d CIFAR-10 datasets. As for the non-i.i.d MNIST and CIFAR-10 datasets, $l=5$.}
  \label{fig:convergence}
\end{figure*}

Thereafter, we change the sampled data size of Per-FedAvg to see how the gap between Per-FedAvg and AutoFL would change with different number of data samples for Per-FedAvg.
Note that when the number of samples increases, the transmit power of UEs in Per-FedAvg is not always $P_{\max}$. Therefore, here we use the \texttt{PowerComputation} algorithm to compute the transmit power with the predifined data sample sizes.
Fig.~\ref{fig:sampled_data_size} shows the results. From Fig.~\ref{fig:sampled_data_size} we observe that, no matter how does the number of data samples change in Per-FedAvg, its overall learning time is smaller than AutoFL. As the sampled data size increase from $5$ to $40$, for the same test accuracy, the average learning time of Per-FedAvg is getting smaller and smaller. This phenomenon is more obvious using the CIFAR-10 dataset. However, when the number of sampled data points is $60$, the learning time performance of Per-FedAvg suddenly deteriorate. We attribute this phenomenon to the reason that the larger sampled data size requires the larger transmit power. And once the transmit power of a certain UE reaches its upper bound, it can not be improved.
Due to insufficient transmit power, the uploads transmitted from this UE fail to arrive the server for global model update.
Therefore, although the number of samples on each UE has increased, the learning time may have become longer and longer.
This result gives us great confidence, that although the original problem (P1) is NP-hard, and AutoFL is designed to approximate the optimal solutions to (P1), the approximate results are also effective in consuming as little time as possible. 

\subsection{Convergence Performance Comparisons} \label{sec:5.3}

Next, we compare convergence performance of AutoFL, Per-FedAvg, FedAvg-1, and FedAvg-2, in terms of training loss and test accuracy. Specifically, the smaller the training loss is, the better the convergence performance is.
On the contrary, as for the test accuracy, the larger, the better.
Fig.~\ref{fig:convergence} shows the convergence performance using the i.i.d MNIST, non-i.i.d MNIST and non-i.i.d CIFAR-10 datasets, respectively.

From Fig.~\ref{fig:convergence}, it is observed that AutoFL outperforms the other three algorithms on the two concerned metrics. This phenomenon is more obvious using the non-i.i.d CIFAR-10 dataset. Besides, AutoFL has the fastest convergence rate, which is consistent with the learning time performance, that AutoFL performs the best.
This result indicates that the $K^*$ in AutoFL is smaller than that in Per-FedAvg and FedAvg. For example, for the non-i.i.d MNIST datasets shown in Fig.~\ref{fig:convergence}, after about $50$ rounds, AutoFL start to converge, while Per-FedAvg starts to converge after about $70$ rounds, the two FedAvg algorithms start to converge after about $100$ rounds.

\subsection{Effect of Network and Dataset Settings} \label{sec:5.4}

\begin{figure*}[t]
  \centering
  \subfloat[The average $K^*$ vs. radius $R$]{
      \includegraphics[width=1.7in]{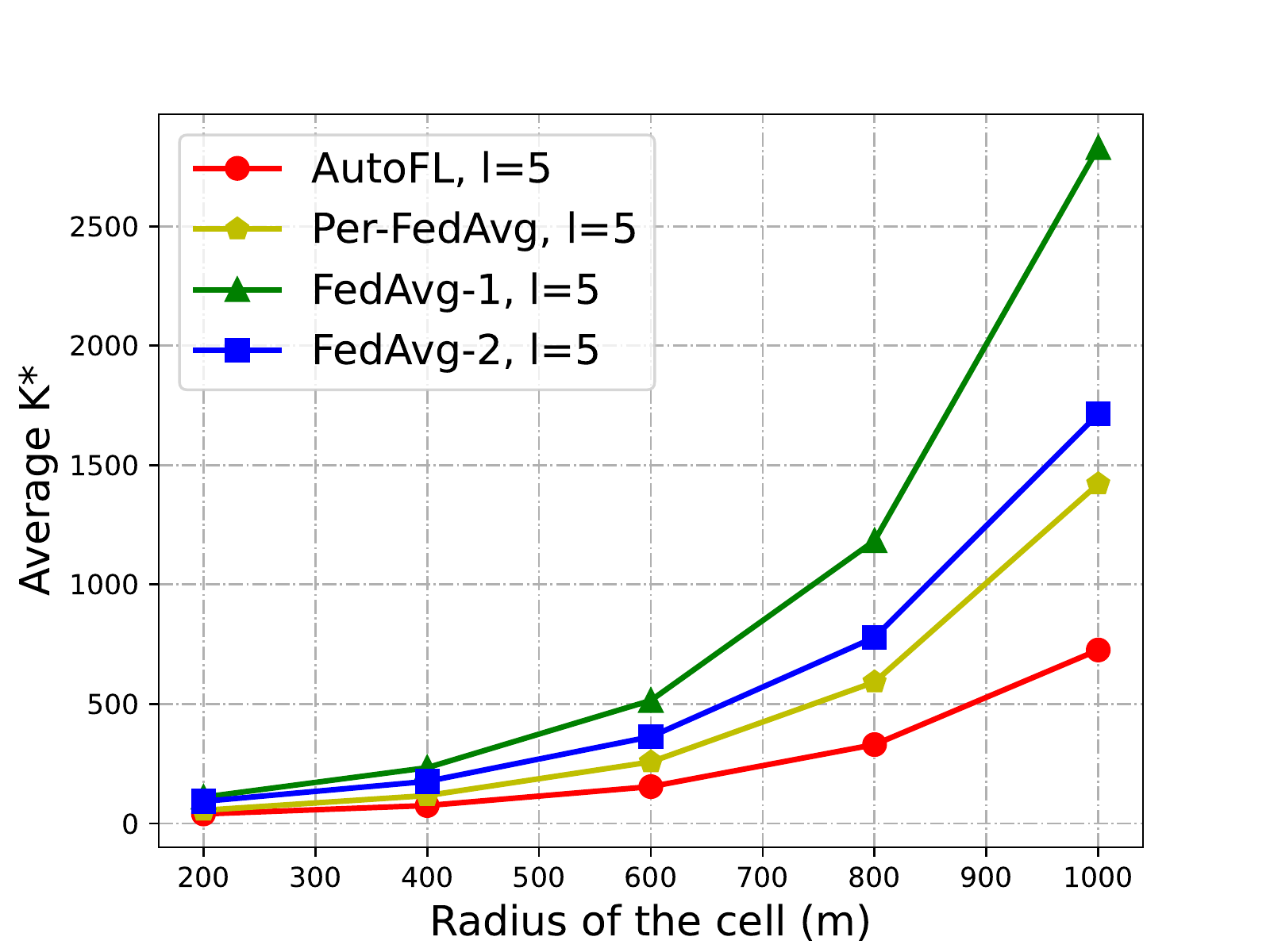}
      \label{fig:distance:subfig:a}}
  \subfloat[The highest achievable accuracy vs. radius $R$]{
      \includegraphics[width=1.7in]{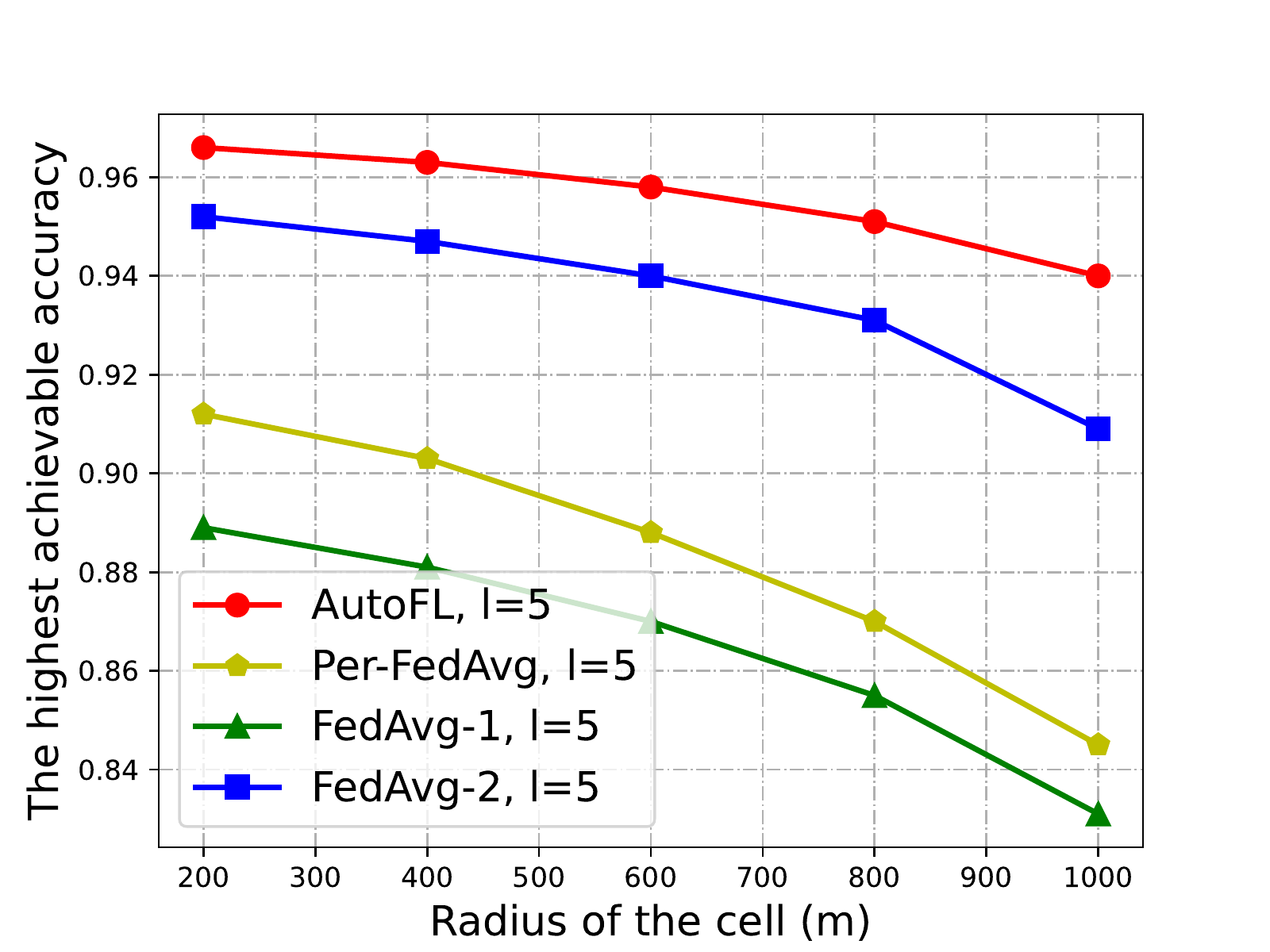}
      \label{fig:distance:subfig:b}}
  \subfloat[The average $K^*$ vs. the non-i.i.d level $l$]{
      \includegraphics[width=1.7in]{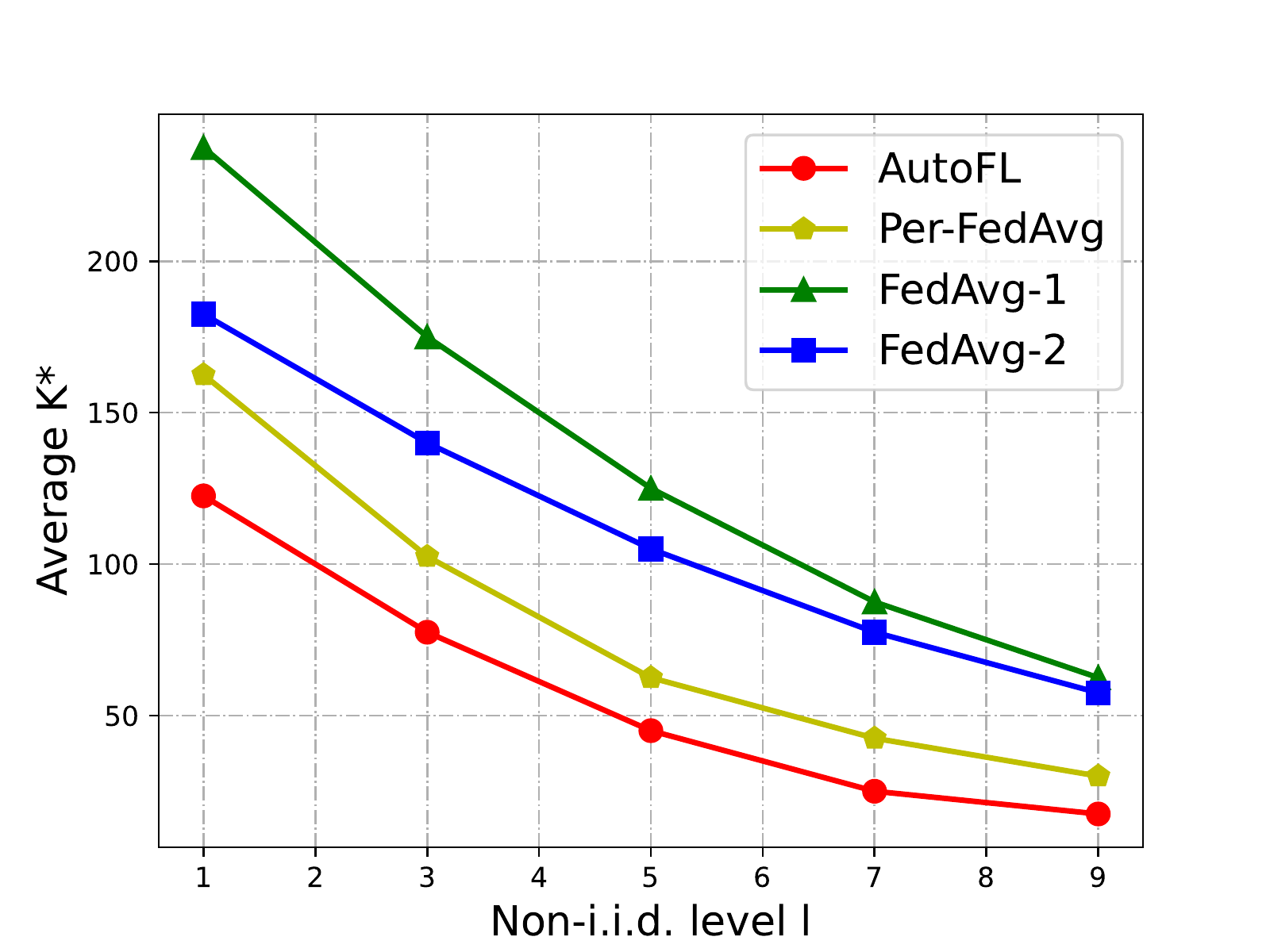}
      \label{fig:labels:subfig:a}}
  \subfloat[The highest achievable accuracy vs. the non-i.i.d level $l$]{
      \includegraphics[width=1.7in]{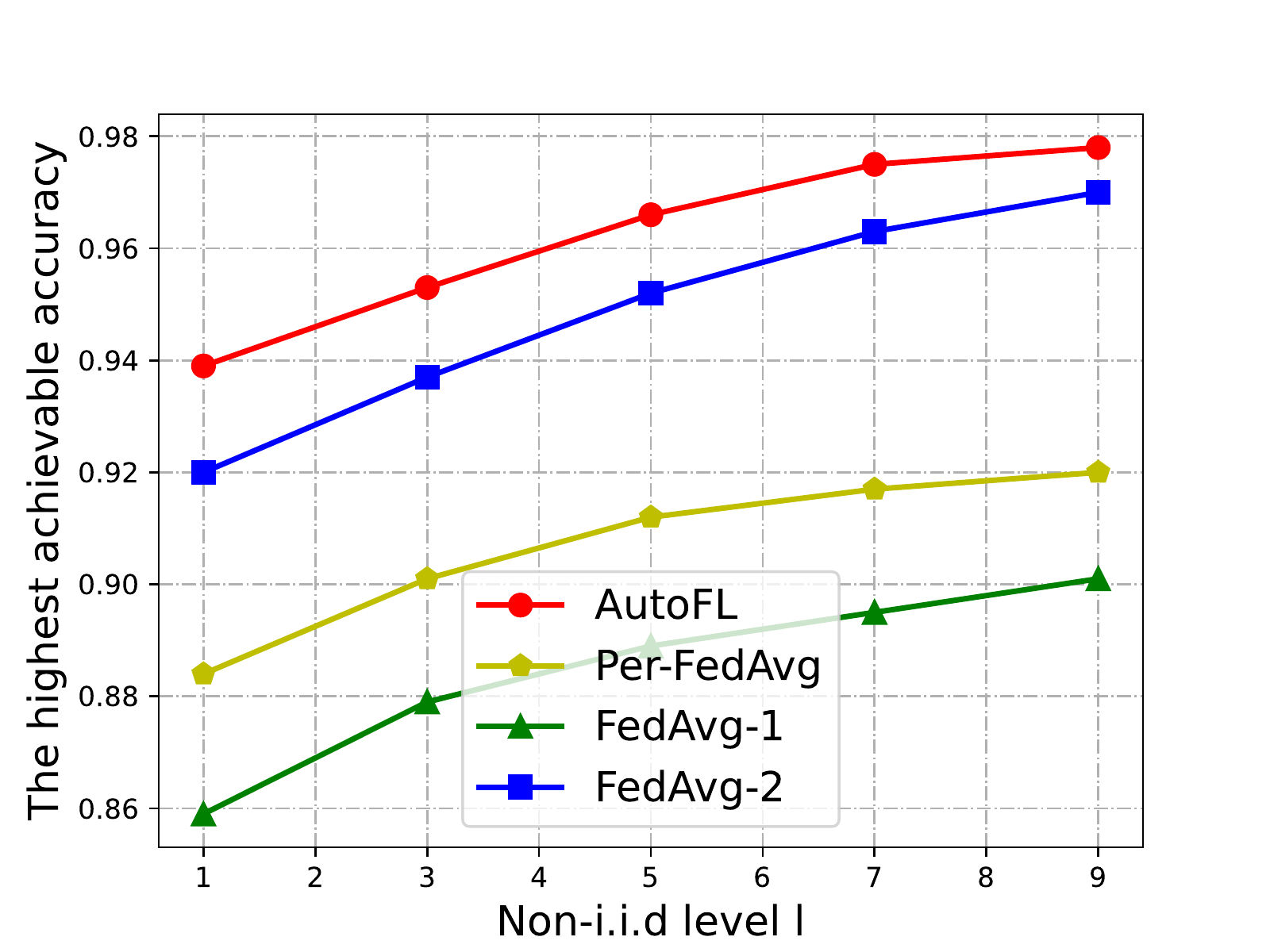}
      \label{fig:labels:subfig:b}}
  \caption{Convergence performance w.r.t. the radius of the cell and the non-i.i.d level $l$ respectively using the MNIST dataset.}
  \label{fig:distance_labels}
\end{figure*}

To understand how different network and dataset settings, such as the radius of the cell and the data heterogeneity, affect the convergence of AutoFL, we conduct a set of experiments on the MNIST dataset.

\begin{itemize}

\item Effect of the radius of the cell $R$: Figs.~\ref{fig:distance:subfig:a} and~\ref{fig:distance:subfig:b} show the average $K^*$ and the highest achievable test accuracy with different radius of the cell $R$. $K^*$ is measured in number of rounds the algorithm starts to converge with test accuracy std $\pm 1\%$.
From Figs.~\ref{fig:distance:subfig:a} and~\ref{fig:distance:subfig:b}, we observe that $K^*$ increases faster and faster with $R$, while the test accuracy decreases faster and faster with $R$. The hidden reason is that, with $R$ increasing, the number of UEs whose local updates can be successfully decoded at the BS decreases. This reduction of UEs' participation in global model update will definitely cause increased $K^*$ and decreased test accuracy. Besides, comparing AutoFL with Per-FedAvg, FedAvg-2 with FedAvg-1, we find that the optimization of transmit power and sampled data size on inividual UEs is beneficial to FL convergence, especially when the wireless environment becomes worse with increasing $R$.

\item Effect of the metric of data heterogeneity $l$: The non-i.i.d level $l$ is used to measure the data heterogeneity. Specifically, the smaller $l$ is, the larger data heterogeneity is. Figs.~\ref{fig:labels:subfig:a} and~\ref{fig:labels:subfig:b} show the average $K^*$ and the highest achievable model accuracy with respect to the non-i.i.d level $l$. We can observe from the figures that as the non-i.i.d level decreases, $K^*$ is decreasing while the test accuracy is increasing. This result is reasonable as the higher degree of data heterogeneity across UEs has more negative impact on the learning process. As expected, Figs.~\ref{fig:labels:subfig:a} and~\ref{fig:labels:subfig:b} also demonstrate that AutoFL can achieve more gains for the more heterogeneous datasets across UEs.

\end{itemize}

\section{Conclusions} \label{sec:6}
In this paper, we have quantified the benefit MAML brings to FL over mobile edge networks.
The quantification is achieved from two aspects: the determination of FL hyperparameters (i.e., sampled data sizes and the number of communication rounds) and the resource allocation (i.e., transmit power) on individual UEs.
In this regard, we have formulated an overall learning time minimization problem, constrained by the model accuracy and energy consumption at individual UEs.
We have solved this optimization problem by firstly analysing the convergence rate of MAML-based FL, which is used to bound the three variables as functions of the model accuracy $\epsilon$.
With these upper bounds, the optimization problem can be decoupled into three sub-problems, each of which considers one of the variables and uses the corresponding upper bound as its constraint.
The first sub-problem guides the optimization of the number of communication rounds. The second and the third sub-problem are computed using the coordinate descent method, to achieve the sampled data size and the transmit power for each UE in each round respectively.
Based on the solutions, we have proposed the AutoFL, a MAML-based FL algorithm that not only quantifies the benefit MAML brings to FL but also maximize such benefit to conduct fast adaptation and convergence over mobile edge networks.
Extensive experimental results have verified that AutoFL outperforms Per-FedAvg and FedAvg, with the fast adaptation and convergence. 

\section*{Appendix}
In order to prove Theorem~\ref{thm:1}, we first introduce an intermediate inference derived from the Lipschitzan gradient assumption. That is, if $f(w)$ is L-Lipschitz continuous, then $\|\nabla f(w) - \nabla f(u)\|\leq L\|w-u\|$ is equivalent to
\begin{equation}\label{equ:appendix_34}
  f(w) \leq f(u) + \nabla f(w)^\top (w-u) + \frac{L}{2} \|w-u\|^2.
\end{equation}

Note that although we assume that each UE only performs one step of gradient descent given the current global model parameter, in the appendix we first consider the general case where $\tau$ ($\tau = 1,2,\dots$) steps of local gradient descents are performed.
Then for each UE $i$, we have
\begin{align}\label{equ:appendix_35}
  \tilde{w}_{k,t}^i = & w_{k,t-1}^i - \alpha \tilde{\nabla} f_i(w_{k,t-1}^i;\mathcal{D}_i^{\text{in}}); \\
  w_{k,t}^i = & w_{k,t-1}^i - \beta (I-\alpha \tilde{\nabla}^2 f_i(w_{k,t-1}^i;\mathcal{D}_i^{\text{o}}))\tilde{\nabla} f_i(\tilde{w}_{k,t}^i;\mathcal{D}_i^{\text{h}}).
\end{align}
where $t=1,\dots,\tau$.
After we find the $\epsilon$-FOSP, we use $\tau = 1$ to obtain the desired result shown in Theorem~\ref{thm:1}. In this case, we introduce the following lemma that has been proved in~\cite{fallah2020personalized} to facilitate our proof.
\begin{lem} \label{lem:4}
  If the conditions in Assumptions 2-4 hold, then for any $\alpha\in[0,1/L]$ and any $t\geq 0$, we have
  \begin{equation} \label{equ:lem_4}
    \mathbb{E}\left[\frac{1}{n}\sum_{i=1}^{n}\|w_{k,t}^i - w_{k,t}\|^2\right] \leq 35 \beta^2 t \tau (2\sigma_F^2 + \gamma_F^2).
  \end{equation}
  where $w_{k,t} = \frac{1}{n} \sum_{i=1}^{n} w_{k,t}^i$.
\end{lem}

Recall that $\bar{w}_{k,t}= \frac{1}{\sum_{i=1}^{n} \mathds{1}\{s_k^i =1, \xi_k^i > \phi\}} \sum_{i=1}^{n}\mathds{1}\{s_k^i =1, \xi_k^i > \phi\} w_{k,t}^i $, from Lemma 1, we know $F(w)$ is $L_F$-Lipschitz continuous, and thus, by (\ref{equ:appendix_34}), we have
\begin{align} \label{equ:appendix_1}
  & F(\bar{w}_{k+1,t+1}) \nonumber \\
  \leq & F(\bar{w}_{k+1,t}) + \nabla F(\bar{w}_{k+1,t+1})^\top (\bar{w}_{k+1,t+1}-\bar{w}_{k+1,t}) \nonumber \\
  & + \frac{L_F}{2} \|\bar{w}_{k+1,t+1} - \bar{w}_{k+1,t}\|^2 \nonumber \\
  \leq & F(\bar{w}_{k+1,t}) - \beta \nabla F(\bar{w}_{k+1,t+1})^\top \left(\sum_{i=1}^{n} U_i \tilde{\nabla}F_i(w_{k+1,t}^i)\right) \nonumber \\
  & + \frac{L_F}{2} \beta^2 \left\|\sum_{i=1}^{n} U_k^i \tilde{\nabla}F_i(w_{k+1,t}^i) \right\|^2,
\end{align}
where the last inequality is obtained given the fact that
\begin{align} \label{equ:appendix_2}
  & \bar{w}_{k+1,t+1} \nonumber \\
  = & \frac{1}{\sum_{i=1}^{n} \mathds{1}\{s_k^i =1, \xi_k^i > \phi\}} \sum_{i=1}^{n}\mathds{1}\{s_k^i =1, \xi_k^i > \phi\} w_{k+1,t+1}^i \nonumber \\
  = & \frac{1}{\sum_{i=1}^{n} \mathds{1}\{s_k^i =1, \xi_k^i > \phi\}} \nonumber \\
  & \sum_{i=1}^{n}\mathds{1}\{s_k^i =1, \xi_k^i > \phi\} \left(w_{k+1,t}^i - \beta \tilde{\nabla}F_i(w_{k+1,t}^i) \right) \nonumber \\
  = & \bar{w}_{k+1,t} - \frac{\beta}{\sum_{i=1}^{n} \mathds{1}\{s_k^i =1, \xi_k^i > \phi\}} \nonumber \\
  & \sum_{i=1}^{n}\mathds{1}\{s_k^i =1, \xi_k^i > \phi\} \tilde{\nabla}F_i(w_{k+1,t}^i) \nonumber \\
  = &  \bar{w}_{k+1,t} - \beta \sum_{i=1}^{n} U_k^i \tilde{\nabla}F_i(w_{k+1,t}^i).
\end{align}
Taking expectation on both sides of (\ref{equ:appendix_1}) yields
\begin{align} \label{equ:appendix_3}
    & \mathbb{E}[F(\bar{w}_{k+1,t+1})] \nonumber \\
    \leq & \mathbb{E}[F(\bar{w}_{k+1,t})]+ \frac{L_F}{2} \beta^2 \mathbb{E}\left[\left\|\sum_{i=1}^{n} U_k^i \tilde{\nabla}F_i(w_{k+1,t}^i) \right\|^2 \right] \nonumber \\
    & - \beta \mathbb{E}\left[\nabla F(\bar{w}_{k+1,t+1})^\top \left(\sum_{i=1}^{n} U_k^i \tilde{\nabla}F_i(w_{k+1,t}^i)\right)\right].
\end{align}
From the above inequality, it is obvious that the key is to bound the term $\sum_{i=1}^{n} U_k^i \tilde{\nabla}F_i(w_{k+1,t}^i)$. Let
\begin{equation} \label{equ:appendix_4}
  \sum_{i=1}^{n} U_k^i \tilde{\nabla}F_i(w_{k+1,t}^i)= X+Y+Z+\sum_{i=1}^{n} U_k^i \nabla F_i(\bar{w}_{k+1,t}),
\end{equation}
where
\begin{align} \label{equ:appendix_5}
  X = & \sum_{i=1}^{n} U_k^i \left(\tilde{\nabla}F_i(w_{k+1,t}^i) - \nabla F_i(w_{k+1,t}^i) \right), \nonumber \\
  Y = & \sum_{i=1}^{n} U_k^i \left(\nabla F_i(w_{k+1,t}^i) - \nabla F_i(w_{k+1,t}) \right), \nonumber \\
  Z = & \sum_{i=1}^{n} U_k^i \left(\nabla F_i(w_{k+1,t}) - \nabla F_i(\bar{w}_{k+1,t}) \right).
\end{align}
Let $\mathcal{F}_{k+1,t}$ be the information up to round $k+1$, local step $t$.
Our next step is to bound the moments of $X$, $Y$, $Z$, conditioning on $\mathcal{F}_{k+1,t}$. Recall the Cauchy-Schwarz inequality
\begin{equation}\label{equ:Cauchy-Schwarz}
  \left\| \sum_{i=1}^{n} a_i b_i\right\|^2 \leq \left(\sum_{i=1}^{n}\|a_i\|^2\right) \left(\sum_{i=1}^{n} \|b_i\|^2\right).
\end{equation}
\begin{itemize}
  \item As for $X$, consider the Cauchy-Schwarz inequality (\ref{equ:Cauchy-Schwarz}) with $a_i = \tilde{\nabla}F_i(w_{k+1,t}^i) - \nabla F_i(w_{k+1,t}^i)$ and $b_i = U_k^i$, we obtain
      \begin{align} \label{equ:appendix_6}
        & \|X\|^2 \nonumber \\ 
        \leq & \left( \sum_{i=1}^{n} U_i^2 \right) \left(\sum_{i=1}^{n}\left\|\tilde{\nabla}F_i(w_{k+1,t}^i) - \nabla F_i(w_{k+1,t}^i)\right\|^2 \right),
      \end{align}
      where $U_i = \max_{k} U_k^i$.
      Hence, by using Lemma 2 along with the tower rule, we have
      \begin{equation} \label{equ:appendix_7}
        \mathbb{E}[\|X\|^2]=\mathbb{E}[\mathbb{E}[\|X\|^2|\mathcal{F}_{k+1,t}]] \leq \left(\sum_{i=1}^{n} U_i^2\right) \sigma_F^2.
      \end{equation}
  \item As for $Y$, consider the Cauchy-Schwarz inequality (\ref{equ:Cauchy-Schwarz}) with $a_i = \nabla F_i(w_{k+1,t}^i) - \nabla F_i(w_{k+1,t})$ and $b_i = U_k^i$, along with the smoothness of $F_i$, we obtain
      \begin{align} \label{equ:appendix_8}
        & \|Y\|^2 \nonumber \\  
        \leq & \left(\sum_{i=1}^{n} U_i^2\right) \left(\sum_{i=1}^{n} \left\| \nabla F_i(w_{k+1,t}^i) - \nabla F_i(w_{k+1,t})\right\|^2 \right) \nonumber \\
        \leq & L_F^2\left(\sum_{i=1}^{n} U_i^2\right) \sum_{i=1}^{n} \|w_{k+1,t}^i - w_{k+1,t}\|^2.
      \end{align}
      Again, taking expectation on both sides of (\ref{equ:appendix_8}) along with the tower rule, we obtain
      \begin{align} \label{equ:appendix_9}
        \mathbb{E}[\|Y\|^2] \leq & L_F^2\left(\sum_{i=1}^{n} U_i^2\right) \mathbb{E}\left[\sum_{i=1}^{n} \| w_{k,t}^i - w_{k,t}\|^2\right] \nonumber \\
        \leq & 35\beta^2L_F^2 n\tau(\tau-1)(2\sigma_F^2 + \gamma_F^2) \left(\sum_{i=1}^{n} U_i^2\right),
      \end{align}
      where the last step is obtained from (\ref{equ:lem_4}) in Lemma~\ref{lem:4} along with the fact that $t\leq \tau-1$.
      \vspace{0.3cm}
  \item As for $Z$, first recall that if we have $n$ numbers $a_1,a_2,\dots,a_n$ with mean $\mu = 1/n \sum_{i=1}^{n} a_i$, and variance $\sigma^2 = 1/n \sum_{i=1}^{n} |a_i - \mu|^2$. If we denote the number of UEs that successfully update their local parameters to the global model as $\delta n$, ($0\leq \delta \leq 1$), according to~\cite{fallah2020personalized}, we have
      \begin{equation} \label{equ:appendix_11}
        \mathbb{E}\left[\left| \sum_{i=1}^{n} U_ia_i-\mu \right|^2\right] = \frac{\sigma^2(1-\delta)}{\delta(n-1)}.
      \end{equation}
      Using this, we have
      \begin{align} \label{equ:appendix_12}
        & \mathbb{E}\left[ \|\bar{w}_{k+1,t} - w_{k+1,t}|\mathcal{F}_{k+1,t}\|^2 \right] \nonumber \\
        \leq & \frac{(1-\delta)\sum_{i=1}^{n}\|w_{k+1,t}^i - w_{k+1,t}\|^2}{\delta(n-1)n}.
      \end{align}
      Therefore, by taking expectation on both sides of (\ref{equ:appendix_12}) along with the tower rule, we have
      \begin{align} \label{equ:appendix_13}
         & \mathbb{E}\left[ \|\bar{w}_{k+1,t} - w_{k+1,t}\|^2 \right] \nonumber \\
         \leq & \frac{35\beta^2(1-\delta)\tau(\tau-1)(2\sigma_F^2 +\gamma_F^2)}{\delta(n-1)},
      \end{align}
      where the inequality is also obtained from (\ref{equ:lem_4}) in Lemma~\ref{lem:4}.
      Next, consider the Cauchy-Schwarz inequality (\ref{equ:Cauchy-Schwarz}) with $a_i = \nabla F_i(w_{k+1,t}) - \nabla F_i(\bar{w}_{k+1,t})$ and $b_i = U_k^i$, we have
      \begin{align} \label{equ:appendix_14}
         & \|Z\|^2 \nonumber \\
         \leq & \left(\sum_{i=1}^{n} U_i^2 \right) \left(\sum_{i=1}^{n}\left\| \nabla F_i(w_{k+1,t}) - \nabla F_i(\bar{w}_{k+1,t}) \right\|^2 \right) \nonumber \\
        \leq & L_F^2 \left(\sum_{i=1}^{n} U_i^2 \right) \sum_{i=1}^{n} \|w_{k+1,t} -\bar{w}_{k+1,t} \|^2.
      \end{align}
      At this point, taking expectation on both sides of (\ref{equ:appendix_14}) along with the use of (\ref{equ:appendix_13}) yields,
      \begin{align} \label{equ:appendix_15}
        & \mathbb{E}[\|Z\|^2] \nonumber \\
        \leq & \frac{35 \beta^2 L_F^2(1-\delta) n \tau(\tau-1)(2\sigma_F^2 +\gamma_F^2)}{\delta(n-1)}\left(\sum_{i=1}^{n} U_i^2 \right).
      \end{align}
\end{itemize}
Now, getting back to (\ref{equ:appendix_3}), we first lower bound the term
\begin{equation} \label{equ:appendix_16}
    \mathbb{E}\left[\nabla F(\bar{w}_{k+1,t+1})^\top \left(\sum_{i=1}^{n} U_i \tilde{\nabla}F_i(w_{k+1,t}^i)\right)\right].
\end{equation}
From (\ref{equ:appendix_4}), we have
\begin{align} \label{equ:appendix_17}
  \mathbb{E}& \left[\nabla F(\bar{w}_{k+1,t+1})^\top \left(\sum_{i=1}^{n} U_i \tilde{\nabla}F_i(w_{k+1,t}^i)\right)\right] \nonumber \\
  = & \mathbb{E}\left[ \nabla F(\bar{w}_{k+1,t+1})^\top \left(X+Y+Z+\sum_{i=1}^{n} U_i \nabla F_i(\bar{w}_{k+1,t})\right) \right]\nonumber \\
  \geq & \mathbb{E}\left[\sum_{i=1}^{n} U_i \|\nabla F(\bar{w}_{k+1,t})\|^2\right] -\|\mathbb{E}\left[ \nabla F(\bar{w}_{k+1,t+1})^\top X\right]\| \nonumber \\
  & - \frac{1}{4} \mathbb{E}[\|\nabla F(\bar{w}_{k+1,t})\|^2] - \mathbb{E}[\|Y+Z\|^2].
\end{align}
This inequality follows the same thought as \cite{fallah2020personalized} by using the fact that
\begin{align} \label{equ:appendix_18}
  & \mathbb{E}[\nabla F(\bar{w}_{k+1,t})^\top (Y+Z)] \nonumber \\
  \leq & \frac{1}{4}\mathbb{E}[\|\nabla F(\bar{w}_{k+1,t})\|^2] + \mathbb{E}[\|Y+Z\|^2].
\end{align}
Meanwhile, the bounds of the terms on the right-hand-side of (\ref{equ:appendix_17}) can be derived similar to what the authors did in \cite{fallah2020personalized}. That is,
\begin{align} \label{equ:appendix_19}
  & \|\mathbb{E}\left[ \nabla F(\bar{w}_{k+1,t+1})^\top X\right]\| \nonumber \\
  \leq & \frac{1}{4}\mathbb{E}[\|\nabla F(\bar{w}_{k+1,t})\|^2] + \mathbb{E}\left[ \left\| \mathbb{E} \left[ X | \mathcal{F}_{k+1,t} \right]\right\|^2\right]\nonumber \\
  \leq & \frac{1}{4}\mathbb{E}[\|\nabla F(\bar{w}_{k+1,t})\|^2] \nonumber \\
  & + \mathbb{E}\left[ \left\|\sum_{i=1}^{n} U_i(\tilde{\nabla} F_i(w_{k+1,t}^i) - \nabla F_i(w_{k+1,t}^i)) \right\|^2\right] \nonumber \\
  \leq & \frac{1}{4}\mathbb{E}[\|\nabla F(\bar{w}_{k+1,t})\|^2] + \left(\sum_{i=1}^{n} U_i^2\right)\frac{4\alpha^2L^2\sigma_G^2}{D}.
\end{align}
Meanwhile, using the Cauchy-Schwarz inequality (\ref{equ:Cauchy-Schwarz}), we have
\begin{align} \label{equ:appendix_20}
  & \mathbb{E}[\|Y+Z\|^2] \nonumber \\
  \leq & 2(\mathbb{E}[\|Y\|^2] + \mathbb{E}[\|Z\|^2]) \nonumber\\
  \leq & 70\beta^2L_F^2n\tau(\tau-1)(2\sigma_F^2 +\gamma_F^2)\left(1+ \frac{1-\delta}{\delta(n-1)}\right)\left(\sum_{i=1}^{n} U_i^2\right) \nonumber\\
  \leq & 140\beta^2L_F^2n\tau(\tau-1)(2\sigma_F^2 +\gamma_F^2)\left(\sum_{i=1}^{n} U_i^2\right).
\end{align}
Combining (\ref{equ:appendix_19}) and (\ref{equ:appendix_20}) yields
\begin{align} \label{equ:appendix_21}
  & \mathbb{E} \left[\nabla F(\bar{w}_{k+1,t+1})^\top \left(\sum_{i=1}^{n} U_i \tilde{\nabla}F_i(w_{k+1,t}^i)\right)\right] \nonumber \\
  \geq & \mathbb{E}\left[ \sum_{i=1}^{n}(U_i-\frac{1}{2}) \|\nabla F(\bar{w}_{k+1,t})\|^2\right] - \left(\sum_{i=1}^{n} U_i^2\right) \frac{4\alpha^2L^2\sigma_G^2}{D} \nonumber \\
  & - 140\beta^2L_F^2n\tau(\tau-1)(2\sigma_F^2 +\gamma_F^2)\left(\sum_{i=1}^{n} U_i^2\right)\nonumber \\
  \geq & \frac{1}{2}\mathbb{E}\left[ \|\nabla F(\bar{w}_{k+1,t})\|^2\right] - \left(\sum_{i=1}^{n} U_i^2\right) \frac{4\alpha^2L^2\sigma_G^2}{D} \nonumber \\
  & - 140\beta^2L_F^2n\tau(\tau-1)(2\sigma_F^2 +\gamma_F^2)\left(\sum_{i=1}^{n} U_i^2\right),
\end{align}
where the last step is derived from the fact that $\sum_{i=1}^n U_i \geq \sum_{i=1}^{n} U_k^i=1$. Next, we characterize an upper bound for the other term in (\ref{equ:appendix_3}):
\begin{equation} \label{equ:appendix_22}
   \mathbb{E}\left[\left\|\sum_{i=1}^{n} U_i \tilde{\nabla}F_i(w_{k+1,t}^i) \right\|^2 \right].
\end{equation}
To do this, we still use the equality (\ref{equ:appendix_4}), that is,
\begin{align} \label{equ:appendix_23}
  & \left\|\sum_{i=1}^{n} U_i \tilde{\nabla}F_i(w_{k+1,t}^i) \right\|^2 \nonumber \\
  \leq &  2\|X+Y+Z\|^2 + 2\left\| \sum_{i=1}^{n} U_i \nabla F_i(\bar{w}_{k+1,t}) \right\|^2 \nonumber \\
  \leq & 4\|X\|^2 + 4 \|Y+Z\|^2 + 2\left\| \sum_{i=1}^{n} U_i \nabla F_i(\bar{w}_{k+1,t}) \right\|^2.
\end{align}
Hence, taking expectations on both sides of (\ref{equ:appendix_23}) along with (\ref{equ:appendix_7}) and (\ref{equ:appendix_20}) we have
\begin{align} \label{equ:appendix_24}
  & \mathbb{E}\left[\left\|\sum_{i=1}^{n} U_i \tilde{\nabla}F_i(w_{k+1,t}^i) \right\|^2 \right] \nonumber \\
  \leq & 2\mathbb{E}\left[ \left\| \sum_{i=1}^{n} U_i \nabla  F_i(\bar{w}_{k+1,t}) \right\|^2 \right] + 4 \sigma_F^2 \left(\sum_{i=1}^{n} U_i^2\right) \nonumber \\
  & + 560\beta^2L_F^2n\tau(\tau-1)(2\sigma_F^2 +\gamma_F^2)\left(\sum_{i=1}^{n} U_i^2\right).
\end{align}
Again, consider the Cauchy-Schwarz inequality with $a_i = \nabla F_i(w_{k+1,t})$ and $b_i = U_i$, we have
\begin{align} \label{equ:appendix_25}
  & \left\| \sum_{i=1}^{n} U_i \nabla F_i(\bar{w}_{k+1,t}) \right\|^2 \nonumber \\
  \leq & \left(\left\| \sum_{i=1}^{n}\nabla F_i(w_{k+1,t}) \right\|^2\right) \left(\sum_{i=1}^{n} U_i^2\right).
\end{align}
By Lemma 3, we have
\begin{equation} \label{equ:appendix_26}
  \mathbb{E}\left[ \|\nabla F_i(\bar{w}_{k+1,t}) - \nabla F(\bar{w}_{k+1,t})\|^2 \right] \leq \gamma_F^2.
\end{equation}
Recall the relationship between the expectation of a random vector $\mathbf{x}$, $\mathbb{E}(\mathbf{x})$ and its variance, $\mathbb{D}(\mathbf{x})$, that $\mathbb{D}(\mathbf{x}) = \mathbb{E}(\mathbf{x}^2) - [\mathbb{E}(\mathbf{x})]^2$. Given that $1/n\sum_{i=1}^{n} \nabla F_i(\bar{w}_{k+1,t}) = \nabla F(\bar{w}_{k+1,t})$, we have
\begin{equation} \label{equ:appendix_27}
  \mathbb{E}\left[\left\|\sum_{i=1}^{n}\nabla F_i(w_{k+1,t})\right\|^2 \right] = \gamma_F^2 + \mathbb{E}[\|\nabla F(\bar{w}_{k+1,t})\|^2].
\end{equation}
Combining (\ref{equ:appendix_24}), (\ref{equ:appendix_25}) and (\ref{equ:appendix_27}) yields
\begin{align} \label{equ:appendix_28}
  & \mathbb{E}\left[\left\|\sum_{i=1}^{n} U_i \tilde{\nabla}F_i(w_{k+1,t}^i) \right\|^2 \right] \nonumber \\
  \leq & 2 \left(\sum_{i=1}^{n} U_i^2\right) \mathbb{E}[\|\nabla F(\bar{w}_{k+1,t})\|^2] \nonumber \\
  & + 560\beta^2L_F^2n\tau(\tau-1)(2\sigma_F^2 +\gamma_F^2)\left(\sum_{i=1}^{n} U_i^2\right) \nonumber \\
  & + 2 \left(\sum_{i=1}^{n} U_i^2\right)\gamma_F^2 + 4 \sigma_F^2 \left(\sum_{i=1}^{n} U_i^2\right).
\end{align}
Substituting (\ref{equ:appendix_21}) and (\ref{equ:appendix_28}) in (\ref{equ:appendix_3}) implies
\begin{align} \label{equ:appendix_29}
  \mathbb{E}& [F(\bar{w}_{k+1,t+1})] \nonumber \\
  \leq & \mathbb{E}[F(\bar{w}_{k+1,t})] - \beta\left[\frac{1}{2} - \beta L_F \sum_{i=1}^{n} U_i^2\right] \mathbb{E}[\|\nabla F(\bar{w}_{k+1,t})\|^2] \nonumber \\
  & + 140(1+2\beta L_F)\beta^3L_F^2n\tau(\tau-1)(2\sigma_F^2 + \gamma_F^2)  \left(\sum_{i=1}^{n} U_i^2\right) \nonumber \\
  & + L_F \beta^2\left(\sum_{i=1}^{n} U_i^2\right)\gamma_F^2 + \beta \left(\sum_{i=1}^{n} U_i^2\right)\frac{4\alpha^2L^2\sigma_G^2}{D} \nonumber \\
  & + 2L_F\beta^2\sigma_F^2 \left(\sum_{i=1}^{n} U_i^2\right)\nonumber \\
  \leq & \mathbb{E}[F(\bar{w}_{k+1,t})] - \frac{\beta}{4}\mathbb{E}[\|\nabla F(\bar{w}_{k+1,t})\|^2] + \beta \sigma_T^2,
\end{align}
where
\begin{align} \label{equ:appendix_30}
  \sigma_T^2 = & 280(\beta L_F)^2 n\tau(\tau-1)(2\sigma_F^2 + \gamma_F^2) \left(\sum_{i=1}^{n} U_i^2\right) \nonumber \\
  & + \beta L_F (2\sigma_F^2 + \gamma_F^2)\left(\sum_{i=1}^{n} U_i^2\right) + \frac{4\alpha^2L^2\sigma_G^2}{D}\left(\sum_{i=1}^{n} U_i^2\right).
\end{align}
The last step of (\ref{equ:appendix_29}) is obtained by $\beta \leq 1/(2\tau L_F)$. Summarizing (\ref{equ:appendix_29}) from $t=0$ to $t=\tau-1$, we have
\begin{align}\label{equ:appendix_31}
  & \mathbb{E}[F(w_{k+1})] \nonumber \\
  \leq & \mathbb{E}[F(w_k)] - \frac{\beta \tau}{4} \left( \frac{1}{\tau} \sum_{t=0}^{\tau-1}\mathbb{E}[\|\nabla F(\bar{w}_{k+1,t})\|^2] \right) + \beta\tau\sigma_T^2,
\end{align}
which is obtained by the fact that $\bar{w}_{k+1,\tau} = w_{k+1}$. Finally, summarizing (\ref{equ:appendix_31}) from $k=0$ to $K-1$, we have
\begin{align} \label{equ:appendix_32}
  & \mathbb{E}[F(w_K)] \leq \mathbb{E}[F(w_0)] \nonumber \\
  & - \frac{\beta \tau K}{4} \left( \frac{1}{\tau K} \sum_{k=0}^{K-1}\sum_{t=0}^{\tau-1}\mathbb{E}[\|\nabla F(\bar{w}_{k+1,t})\|^2] \right) + \beta\tau K\sigma_T^2.
\end{align}
As a result, we have
\begin{align}\label{equ:appendix_33}
  & \frac{1}{\tau K} \sum_{k=0}^{K-1}\sum_{t=0}^{\tau-1}\mathbb{E}[\|\nabla F(\bar{w}_{k+1,t})\|^2] \nonumber \\
  \leq & \frac{4}{\beta \tau K}(F(w_0) - \mathbb{E}[(F(w_K))] + \beta \tau K \sigma_T^2) \nonumber \\
  \leq & \frac{4(F(w_0) - F^*)}{\beta\tau K} + 4 \sigma_T^2.
\end{align}
Given that we only consider one step of local SGD in this paper, we use $\tau = 1$, and then the desired result is obtained.

{
\footnotesize
\bibliographystyle{IEEEtran}
\bibliography{IEEEabrv,IEEEexample}
}

\end{document}